\documentclass[a4paper,fleqn]{cas-dc}

\usepackage[authoryear]{natbib}

\usepackage{booktabs}

\usepackage[printonlyused,withpage,nolist,nohyperlinks]{acronym}
\usepackage{color,soul}
\usepackage{siunitx}

\usepackage{amsmath}

\usepackage{graphicx}
\usepackage{float}
\usepackage{hyperref}

\begin{document}
\let\WriteBookmarks\relax
\def\floatpagepagefraction{1}
\def\textpagefraction{.001}

\shorttitle{Automatic Scale Estimation of Structure from Motion based 3D Models using Laser Scalers}
\shortauthors{K. Isteni\v{c} et~al.}

\title [mode = title]{Automatic Scale Estimation of Structure from Motion based 3D Models using Laser Scalers}

\author[1,2]{Klemen Isteni\v{c}}[orcid=0000-0003-1911-1483]
\cormark[1]
\ead{klemen.istenic@gmail.com}

\author[1]{Nuno Gracias}[orcid=0000-0002-4675-9595]
\ead{ngracias@silver.udg.edu}

\author[3]{Aur\'elien Arnaubec}[]
\ead{aurelien.arnaubec@ifremer.fr}

\author[4]{Javier Escart\'in}[orcid=0000-0002-3416-6856]
\ead{escartin@ipgp.fr}

\author[1]{Rafael Garcia}[orcid=0000-0002-1681-6229]
\ead{rafael.garcia@udg.edu}

\address[1]{Underwater Robotics Research Center (CIRS), Computer Vision and Robotics Institute (VICOROB), University of Girona, Edifici P-IV, Campus de Montilivi, 17071 Girona, Spain}

\address[2]{Coronis Computing, S.L., Science and Technological Park of UdG, Carrer Pic de Peguera, 15, 17003 Girona, Spain}
\address[3]{IFREMER, Ctr Mediterranee, Unit\'e Syst.~Marins, CS 20330, F-83507 La Seyne Sur Mer, France}
\address[4]{Universit\'e de Paris, Institut de Physique du Globe de Paris, CNRS, F-75005, France}

\cortext[cor1]{Corresponding author}

\sloppy

\begin{abstract}

Recent advances in structure-from-motion techniques are enabling many scientific fields to benefit from the routine creation of detailed 3D models. However, for a large number of applications, only a single camera is available for the image acquisition, due to cost or space constraints in the survey platforms. Monocular structure-from-motion raises the issue of properly estimating the scale of the 3D models, in order to later use those models for metrology. The scale can be determined from the presence of visible objects of known dimensions, or from information on the magnitude of the camera motion provided by other sensors, such as GPS. 

This paper addresses the problem of accurately scaling 3D models created from monocular cameras in GPS-denied environments, such as in underwater applications. Motivated by the common availability of underwater laser scalers, we present two novel approaches which are suitable for different laser scaler configurations. A fully-calibrated method enables the use of arbitrary laser setups, while a partially-calibrated method reduces the need for calibration by only assuming parallelism on the laser beams, with no constraints on the camera. The proposed methods have several advantages with respect to the existing methods. By using the known geometry of the scene expressed by the 3D model, along with some parameters of the laser scaler geometry, the need for laser alignment with the optical axis of the camera is removed. Furthermore, the extremely error-prone manual identification of image points on the 3D model, currently required in image-scaling methods, is eliminated as well.

The performance of the methods and their applicability was evaluated on both data generated from a realistic 3D model and data collected during an oceanographic cruise in 2017. Three separate laser configurations have been tested, encompassing nearly all possible laser setups, to evaluate the effects of terrain roughness, noise, camera perspective angle and camera-scene distance on the final estimates of scale. In the real scenario, the computation of $6$ independent model scale estimates using our fully-calibrated approach, produced values with standard deviation of $0.3\%$. By comparing the values to the only possible method usable for this dataset, we showed that the consistency of scales obtained for individual lasers is much higher for our approach ($0.6\%$ compared to $4\%$).
\end{abstract}



\begin{keywords}
Structure-from-Motion \sep Underwater 3D Reconstruction \sep Photogrammetry \sep Laser Scalers
\end{keywords}

\maketitle

\begin{acronym}

\acro{2D}{2-dimensional}
\acro{2.5D}{2.5-dimensional}
\acro{3D}{3-dimensional}
\acro{AD*}{anytime dynamic A*}
\acro{AGP}{art gallery problem}
\acro{ASV}{autonomous surface vehicle}
\acro{AUV}{au\-to\-no\-mous un\-der\-wa\-ter vehicle}
\acrodefplural{AUV}[AUVs]{autonomous underwater vehicles}
\acro{BMS}{battery management system}
\acro{C-Space}{configuration space}
\acro{CIRS}{underwater robotics research center}
\acro{CL-RRT}{closed-loop rapidly-exploring random tree}
\acro{COLA2}{component oriented layer-based architecture for autonomy}
\acro{CPF}{cooperative path following}
\acro{CPP}{coverage path planning}
\acro{DFS}{depth-first search}
\acro{DOF}{degree of freedom}
\acrodefplural{DOF}[DOFs]{degrees of freedom}
\acro{DVL}{Doppler Velocity Log}
\acro{EKF}{extended Kalman filter}
\acro{EST}{expansive-spaces tree}
\acro{FB}{frontier-based}
\acro{FM}{fast marching}
\acro{FOV}{field of view}
\acrodefplural{FOV}[FOVs]{fields of view}
\acro{GA}{genetic algorithm}
\acro{GNC}{guidance, navigation and control}
\acro{GPS}{global positioning system}
\acro{IMU}{inertial measurement unit}
\acro{INS}{inertial navigation system}
\acro{KF}{Kalman filter}
\acro{LKH}{Lin-Kernighan-Helsgaun heuristic}
\acro{LOS}{line of sight}
\acro{LS}{least squares}
\acro{MSV}{manned submersible vehicle}
\acrodefplural{MSV}[MSVs]{manned submersible vehicles}
\acro{NBV}{next-best-view}
\acro{NED}{north-east-down}
\acro{NOAA}{national oceanic and atmospheric administration of United States}
\acro{OMPL}{open motion planning library}
\acro{PDN}{perception-driven navigation}
\acro{PID}{proportional-integral-derivative}
\acro{PMP}{partial motion planner}
\acro{PRM}{probabilistic roadmap}
\acro{ROS}{robot operating system}
\acro{ROV}{remotely operated vehicle}
\acrodefplural{ROV}[ROVs]{remotely operated vehicles}
\acro{RIG}{rapidly-exploring information gathering}
\acro{RRG}{rapidly-exploring random graph}
\acro{RRT}{rapidly-exploring random tree}
\acro{RRT*}{asymptotic optimal rapidly-exploring random tree}
\acro{ROI}{region of interest}
\acro{RPP}{randomized path planner}
\acro{SAS}{synthetic aperture sonar}
\acro{SLAM}{Simultaneous Localization And Mapping}
\acro{STOMP}{stochastic trajectory optimization for motion planning}
\acro{T-RRT}{transition-based rapidly-exploring random tree}
\acro{TSP}{traveling salesman problem}
\acro{UAV}{unmanned aerial vehicle}
\acrodefplural{UAV}[UAVs]{unmanned aerial vehicles}
\acro{UdG}{university of Girona}
\acro{UGV}{unmanned ground vehicle}
\acrodefplural{UGV}[UGVs]{unmanned ground vehicles}
\acro{UUV}{unmanned underwater vehicle}
\acro{UV}{underwater vehicle}
\acro{UWSim}{underwater simulator}
\acro{VICOROB}{computer vision and robotics group}
\acro{VP}{view planning}
\acro{WP}{waypoint}
\acro{YAML}{YAML ain't markup language}
\acro{SfM}{Stru\-cture from Mo\-tion}
\acro{BA}{Bundle Adjustment}
\acro{RANSAC}{RANdom SAmple Consensus}
\acro{AC-RANSAC}{A Contrario Ransac}
\acro{NLS}{non-linear least squares}
\acro{PnP}{Perspective-n-Point}
\acro{P3P}{Perspective-3-Point}
\acro{DLT}{Direct Linear Transform}
\acro{GPU}{graphics processing unit}
\acro{MC}{Monte Carlo}
\acro{LM}{Levenberg-Marquardt}
\acro{USBL}{Ultra-Short BaseLine}
\acro{GNSS}{Global Navigation Satellite System}
\acro{GCP}{ground control point}
\acro{GSD}{ground sample distance}
\end{acronym}

\section{Introduction}
An increasing number of remote sensing applications are emerging, relying on photogrammetry to obtain reliable geometric information about the environment. These optical-based reconstruction procedures, generally based on the 
\ac{SfM} approach, have gained significant popularity due to multiple factors. The improvements in both speed and robustness of many image processing techniques~\citep{snavely2008modeling,remondino2008turning,agarwal2009building,triggs1999bundle} together with increased computational capabilities of commonly available processing hardware, enable nowadays nearly black-box type of data pro\-cess\-ing, where there is little to no need for user intervention. The abundance of low cost cameras that can easily be mounted on a variety of vehicles, or used hand-held, has further spearheaded the widespread of these techniques in a variety of fields (e.g.,~\citealp{wallace2016assessment,javernick2014modeling,anderson2013light,mathews2013}). 

\begin{figure}
\centering
\includegraphics[width=8cm]{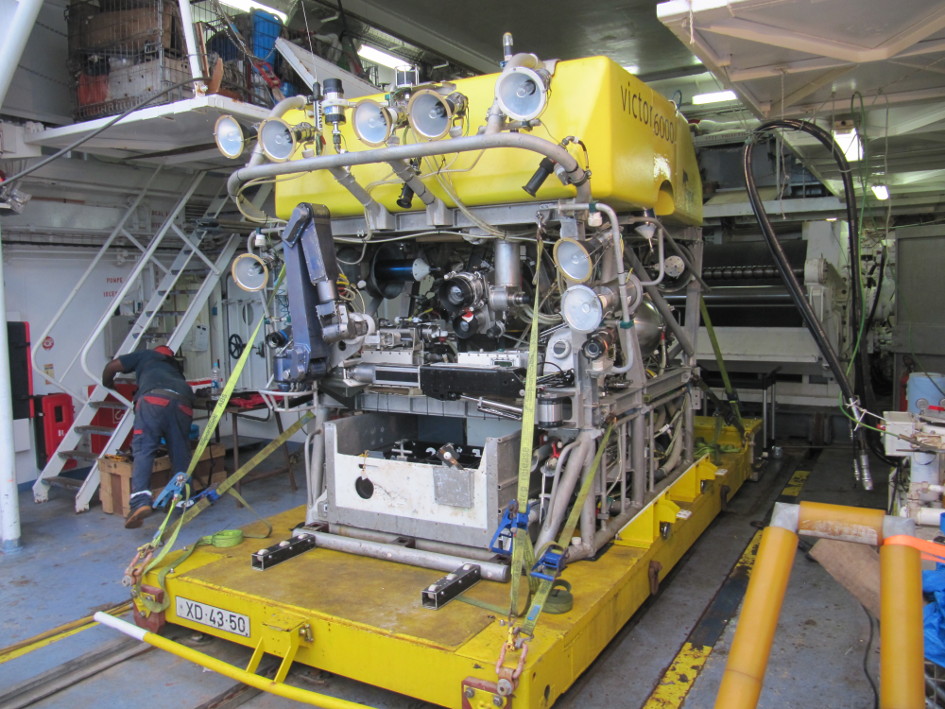}
\caption{ROV VICTOR 6000 (IFREMER), used among other, in the SUBSAINTES 2017 cruise  (doi:10.17600/17001000).}
\label{fig:res_victor}
\end{figure}

Concurrently, the field of underwater photogrammetry has also grown considerably with the availability of underwater vehicles. Whereas traditional aerial and terrestrial vehicles are increasingly equipped with single or multi-camera set-ups (e.g.,~stereo cameras, multi-camera systems), most underwater \acp{ROV} and \acp{AUV} that are nowadays used in science missions (e.g., VICTOR 6000 from IFREMER depicted in Fig.~\ref{fig:res_victor}) have limited optical sensing capabilities. Common optical systems consist of a single main camera used by the \ac{ROV}-pilot or, in case of larger workclass \acp{ROV}, also of additional cameras for maneuvering. As these are typically unsynchronized and have non-overlapping fields-of-view, they are not suited for stereo image processing. Nonetheless, the ability to produce accurate \ac{3D} models from monocular cameras despite the unfavorable properties of the water medium (i.e.,~light attenuation and scattering, among other effects) has given scientists unprecedented access to the underwater environment and its ecosystems, from shallow waters~\citep{pizarro2017simple,storlazzi2016end,rossi2019detecting} to the deep ocean \citep{bingham2010robotic,escartin2016first,bodenmann2017generation}.

Performing \ac{SfM} based reconstruction using single camera imagery has an important limitation as it precludes obtaining a metric scale of the resulting model. The image formation process of projecting the \ac{3D} world onto \ac{2D} image planes causes the loss of a dimension. When performing the reconstruction, this results in scale ambiguity, i.e.~the estimated parameters of 3D structure and camera trajectory can be multiplied with an arbitrary factor and still give rise to the same image observations~\citep{lourakis2013accurate,zisserman2004multiple}. This also precludes or at least limits the possibility to conduct quantitative measurements based on geometric parameters (e.g., distances, areas, angles, etc.) obtained from the models. To resolve the ambiguity, a general trend in sub-aerial problems is to fuse the image measurements with other sensors  (e.g.,~\ac{INS}~\citep{spaenlehauer2017loosely,ji2015visual} and \ac{GNSS}\citep{soloviev2010integration,mian2016accuracy,Forlani2018quality} ) or using \acp{GCP}\citep{james2014mitigating, eltner2015analysis,mertes2017rapid}. These geometric control points are extremely hard, if not impossible, to establish underwater, while the absorption of electromagnetic waves in water prevents the use of GPS. Hence the scale is normally disambiguated either using \ac{INS}~\citep{sedlazeck2009kiel,pizarro2009large,campos2016underwater} or through the introduction of known distances between points in the scene~\citep{garcia2011high}. It is worth noting that reliable displacement information may not be available in smaller \acp{ROV}, since this normally requires a dedicated \ac{INS} complemented with a \ac{DVL}. Given there are rarely any known measurements readily available in real underwater scenarios, the scale is therefore often determined by placing objects with known dimensions (e.g.,~scaling cube~\citep{cocito2003bio}, locknuts~\citep{kalacska2018freshwater}, graduated bars~\citep{neyer2018monitoring}, etc.) into the scene. While such approach does not require any additional equipment (with the exception of auxiliary objects), it does however involve their transport and placement, which can be challenging in deep-sea environments. 

Alternatively, the distance between known points on the model can also be established from the projections of laser beams with known geometry~\citep{robert2017new, bergmann2011mega, tusting1992laser}. The use of laser scalers to provide an absolute size reference in photographs is one of its most widespread uses~\citep{tusting1992laser,tusting1993improved}. Their initial use dates back to the late 1980s~\citep{tusting1986noncon,caimi1987application}. To compensate the lack of knowledge about the scene and camera-scene distance, the methods require a perfect alignment of parallel lasers with the camera, planarity of the scene surface and perpendicularity between the camera and the scene. Comparing the spacing between two laser spots on the image and the known beam spacing, any measurement in the plane of the lasers, regardless of the camera-to-scene range, should be correctly estimated. 

Seen as the most restrictive requirement, the necessity of perpendicularity between the optical axis of the camera and the scene has been addressed in various approaches with the introduction of additional lasers and sensors. Wakefield et al.~\cite{wakefield1987canadian} first introduced the idea of perspective grids to enable oblique camera views. Although being a progress, the method imposed additional constraints on the camera-scene distance (altitude) and fixed inclination angle. 

To provide additional information about the camera-scene relationship, more lasers have also been added to the systems. A configuration consisting of three lasers, two aligned with the optical axis of the camera and a third laser oriented at an angle, has been described by Tusting and Davis~\citep{caimi1993advanced}. It enables the estimation of range and size of objects from direct scaling of the position of the light spots on the image. An underwater photogrammetric system using several sensors to provide precision navigation for benthic surveys is described in \cite{kocak2002laser,kocak2004remote}. One of them, the ring laser gyroscope, made for measuring pitch/roll motions is integrated in a custom software package which establishes the scale reference from the projections of the three beam laser system. To enable the measurement of distance between any two points on the image, Pilgrim et al.~\citep{pilgrim2000rov} presented a multi-laser approach. It gains the information about the camera's inclination angle and distance to the scene by using four parallel lasers positioned equidistant from the camera center together with a fifth laser set at an angle either parallel to the bottom or a side pair, similar to the three-beam approach. The method works under the assumption of scene flatness and the restraint of the camera in either pan or tilt planes with respect to the sea bottom. A more versatile method capable of determining an arbitrary tilt of a surface was presented by Davis and Tusting~\citep{davis1991quantitative} which requires four parallel lasers aligned with the optical axis of the camera.

Due to the lack of a better approach, image scaling methods are still commonly used for scaling 3D models, and therefore require not only for the images observing the projections of lasers to be acquired in flat areas of the scene, but also complex laser alignment with the optical axis of the camera. Depending on the circumstances (multiple dives with mounting and dismounting of equipment), these strict rigidity constraints can be nearly impossible to maintain in real scientific cruises where camera might not be rigidly coupled with the laser rig, among other problems. As accurate geometrical information would entail repetitive calibration procedures, it significantly limits its usability. Furthermore, given that the image scaling techniques only provide the estimated distance between points on an image, this information is not directly related to the model itself. In order to scale any model, a separate identification of these laser points has to be done on the model itself. As the identification of image points on the model is done manually, it is extremely error prone and time consuming.


The main goal of this paper is to present two novel automatic approaches to solve the scaling problem for \ac{SfM} based 3D models, using commonly available laser scalers. The image information is exploited beyond the automatic location of laser spots, compensating for known geometry of the laser scalers. The need for laser alignment with the optical axis is thus abolished together with the manual identification of 3D points on the model, which is prone to errors. 

Each of the two proposed methods (i.e., fully- and partially -calibrated) is suitable for a different laser scaler configuration. While the fully-calibrated approach enables an arbitrary laser setup, the required rigidity between the lasers and the camera can be extremely limiting in real scenarios. To overcome this, we also present an alternative approach in which the relation of the lasers to the camera is significantly reduced at the cost of requiring the lasers to be parallel among them (not necessarily with the optical axis). As fully-calibrated method utilizes a fully-determined laser geometry, it is able to estimate the scale using a single laser while the partial method requires a laser pair. Any additional laser measurements are used to further reduce potential effect of noisy laser spot detections. These methods are considered universal, as they can be applied to standard imagery acquisitions, and are not not linked to data acquired with specific sensors or hardware (e.g., stereo cameras). Hence, it is possible to process legacy data from previous missions acquired using different vehicles and imaging systems. 

The results of our methods are validated using a 3D model constructed using real underwater data and compare them to results which would have been obtained using an image scaling method supporting arbitrary tilt of the surface~\citep{davis1991quantitative}. The effects of noise, camera perspective angle and camera-scene distance on our process and final estimates of scale are further analyzed. Finally, the results of using our method to scale a model reconstructed from data acquired during the SUBSAINTES 2017 cruise  (doi:10.17600/17001000)~\citep{escartin2017} are presented.

\section{Scaling of SfM-based 3D Models}
Optical-based \ac{3D} models are produced using a set of images through a series of sequential steps. A sparse set of 3D points representing the general 3D geometry of the scene can be obtained by exploiting multiple projections of the same 3D point in overlapping images through the equations of projective geometry~\citep{zisserman2004multiple}. By extracting salient features and matching them across the image set, the 3D locations of these points (the structure) are estimated together with the camera parameters (the motion) through a technique called Structure from Motion (SfM). An accurate high-detailed description of the model is subsequently obtained through an efficient multi-view stereo densification process. This is followed by an estimation of a surface from the obtained unorganized noisy set of 3D points (point cloud). The final photo-realistic 3D model is obtained by finding a consistent high-quality texture by seamlessly mapping input images to a high-resolution triangle representation of the surface. If the imagery used in the process was acquired using one or more unsynchronized cameras, and no other auxiliary data is used, it is impossible to determine the correct scale of the model. Such result can be visually pleasing but cannot be used for further scientific purposes where knowledge of the distances, areas and volumes is required. Therefore, a scale estimation step is vital in the reconstruction for scientific purposes. 

\begin{figure*}
	\centering
	\includegraphics[height=7cm]{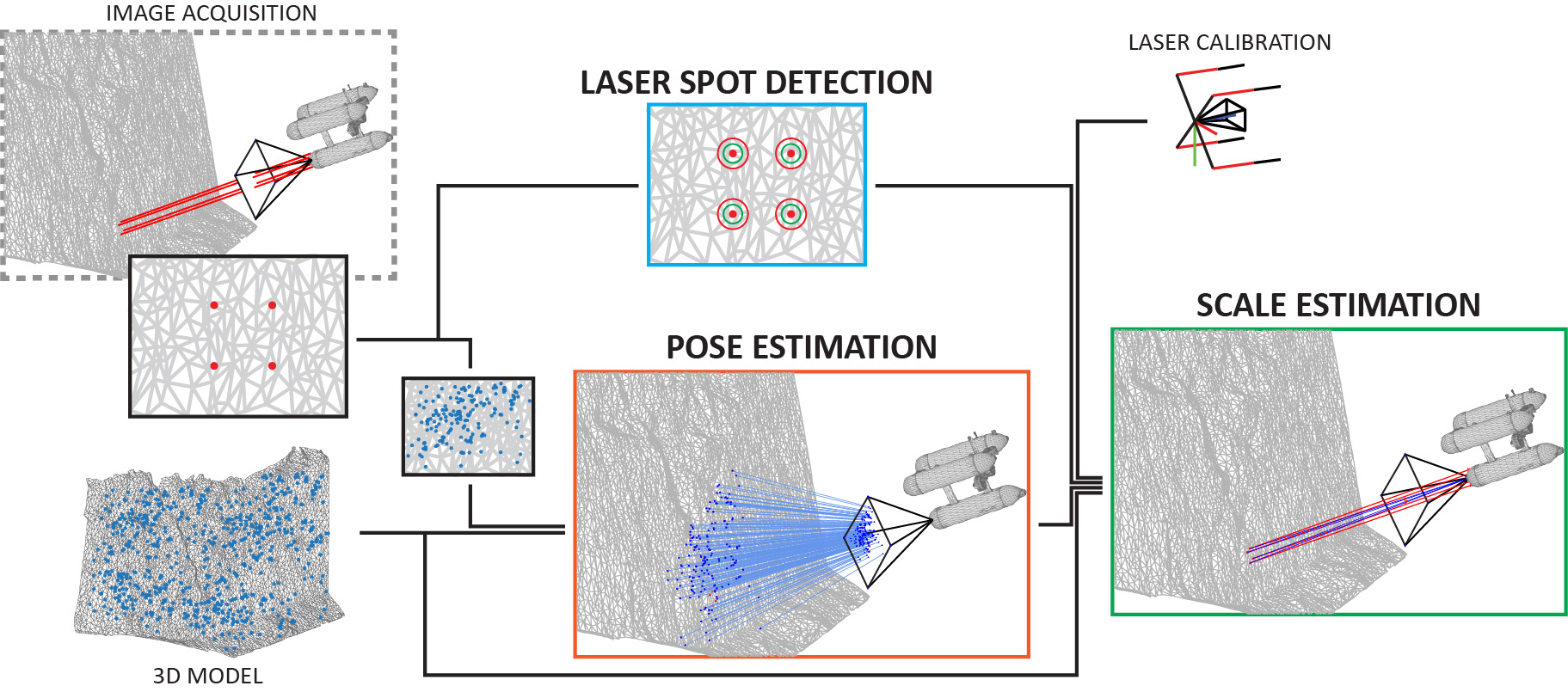}
	\caption{Flowchart of the scale estimation process depicting three crucial steps in scale estimation: laser spot detection, pose estimation, and scale estimation.}
	\label{fig:eval_scheme}
\end{figure*}

Nowadays, the most common uses of laser scalers are for image scaling and are based on multi-laser approaches introduced by Pilgrim et al.~\citep{pilgrim2000rov} and Davis and Tusting~\citep{davis1991quantitative}. The requirements associated with these methods, i.e.~laser alignment with the optical axis and manual identification of the image points on the 3D models, while once reasonable, are becoming constricting in increasing number of occasions in which data for photogrammetry can be collected. 

In this section, we present two novel methods for scale estimation, namely fully calibrated method (FCM) and partially calibrated method (PCM), suitable for different laser scaler configurations and scenarios. Both methods, based on computer vision techniques of image localization and ray casting, exploit the information acquired with an optical image in which the intersection of lasers with the scene (laser spots) are visible. Both methods consist of three main steps, as depicted in Fig.~\ref{fig:eval_scheme}. The two initial steps are identical in both methods. First, a laser detection method is required to determine the  locations of laser spots on an image. Secondly, the pose of the camera (wrt.~the 3D model), at the moment at which the image was acquired, is estimated through a feature-based localization process. These estimations are used in the third step, which differs between methods and depends on available laser configuration information. The scale of the model is computed after determining the 3D position of laser beams intersecting with the scene. 

It is worth noting that our approaches are independent of the method used for detecting laser spots on the image. Laser spots can be selected either manually, through a simple method (e.g., color thresholding) or even with a more complex approach (e.g., machine learning~\citep{rzhanov2005uvsd}).

\subsection{Measuring device}

\begin{figure}
	\centering
	\begin{tabular}{@{}cc}
		\includegraphics[height=3.5cm]{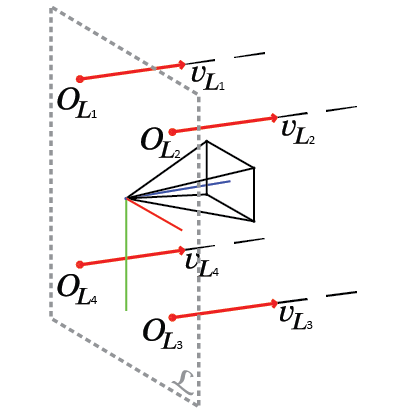} &
		\includegraphics[height=3.5cm]{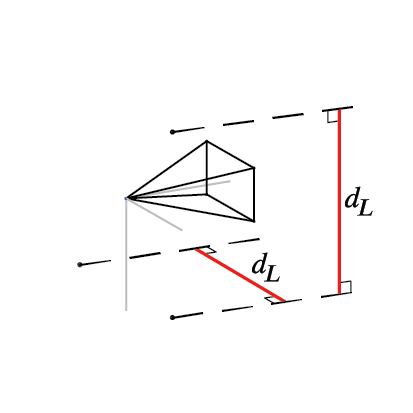} \\
		(a) & (b) 
	\end{tabular}
	\caption{a) Fully- and b) partially-calibrated measuring device (optical camera and separate lasers) with the required information marked in red.}
	\label{fig:deviceConfig}
\end{figure}

The measuring setup required consists of two devices commonly used in underwater surveying using ROVs and AUVs: A laser scaler, which can contain a variable number of lasers, and a monocular optical camera. If the laser geometry (origins $O_{\!L}$ and directions $v_{\!L}$) with respect to the optical axis of the camera are known, the setup is considered fully calibrated (Fig.~\ref{fig:deviceConfig}a). The origins are defined as points on a plane $\mathcal{L}$, which is perpendicular to the optical axis of the camera and contains the optical center, while the directions are unit vectors expressed wrt.~the camera's optical axis. These geometric relations can be easily obtained through a calibration procedure, in which the camera captures images with clearly visible laser-surface intersections, and with a distance to the camera that is either known or that can be easily computed. Each intersection is then represented by a 3D point in the camera frame, and beams directions can be estimated by finding best fitting lines. Subsequently, computing the point of intersection between the fitted lines and plane  $\mathcal{L}$ reveals the laser origins. 

Depending on the circumstances (e.g.,~multiple dives involving mounting and dismounting of equipment with associated misalignements), the strict rigidity constraints between the lasers and the camera is very difficult to maintain, especially if the camera and laser scaler are not rigidly attached. As any change would thus entail a new calibration procedure, which is not systematically done and may be unfeasible, we also present an alternative approach, in which laser pairs have to be parallel with the sole condition of the camera being equidistant to their origins (Fig.~\ref{fig:deviceConfig}b). As there is no requirement of parallelism between the laser beams and the optical axis of the camera, this partially calibrated approach permits alterations between the camera and laser scaler making it more suitable for scenarios with multiple mounting and dismounting operations, or situations in which accurate calibration procedure is not possible or unavailable. These relaxed constraints render the system more usable in practice.

\subsection{Pose Estimation}
\label{sec:pose_estimation}
The scale estimation process requires the knowledge of the camera pose $\boldsymbol{P}\! =\!\left[\boldsymbol{R}^T\!\mid\!-\boldsymbol{R}^T\boldsymbol{t}\right] \in \mathbf{SE} (3)$ defined as projection from world to camera frame at the moment the image was taken. As these images contain lasers spots, they do not reflect the real state of the environment and are as such considered undesirable in the 3D reconstruction process. Therefore, in order to estimate their poses (wrt.~the 3D model), a feature-based image localization method is used.  

Salient 2D features extracted from the image, are matched with a full set of features associated with the model's sparse set of 3D points. Feature detection and matching procedures can be adjusted for each specific dataset, and do not influence the scale estimation process, as long as it is possible to produce successful pairs of 3D-2D observations ($\mathcal{F} = \{X_{\!k},x_{\!j}\}$). Such matches are then exploited to obtain an initial estimate of camera extrinsic parameters $\boldsymbol{P}$ (and possible camera intrinsics $\boldsymbol{K}$). In cases in which the camera is calibrated, the solution is obtained by solving a minimal case ($n\!=\!3$) of the \ac{PnP} problem~\citep{ke2017efficient}, while alternatively a \ac{DLT}~\citep{zisserman2004multiple} algorithm can be used. As feature observations are noisy and might contain outliers, the process is done in conjunction with a robust estimation method \ac{AC-RANSAC}~\citep{moisan2012automatic}. Initial parameter values are subsequently refined through a non-linear optimization. Using \ac{BA} the the re-projection error of known (and fixed) 3D points and their 2D observation is minimized:

\begin{equation}
\min\limits_{P,K}
\sum_{\mathcal{F}} \displaystyle \big\Vert  x_{\!j} - \text{proj}(\boldsymbol{K},\boldsymbol{P},\mathbf{X}_{k})\big\Vert^{2} \:.
\label{eq:loc_reproj_error}
\end{equation}

\subsection{Scale estimation}

In our approaches the scale of a 3D model is obtained as the ratio between a known quantity $m$ and its model based estimate $\hat{m}$:

\begin{equation}
s = \frac{m}{\hat{m}}\:.
\label{eq:scale}
\end{equation}

Using the location of recorded and detected laser spots $x_{\!L}$ and previously estimated parameters of the camera $\{\boldsymbol{K}, \boldsymbol{P}\}$, it is possible to predict the geometry of the laser scaler which produced the recorded results. Given that the prediction is based on the 3D model, it is directly affected by the scale of the model and can therefore be used to determine it. Depending on the availability of information about the geometry of the lasers and the camera, we can either use the distance between the laser origins and camera's optical center (FCM) or the perpendicular distance between the two parallel beams (PCM). 

\subsubsection{Fully calibrated method}

As complete laser geometry (origins $O_{\!L}$ and directions $v_{\!L}$) is known, the position from where the lasers had to be emitted $\hat{O}_{\!L}$ in order to produce the observed result can be determined regardless of potential non-parallelism between the lasers. The position of origin of each laser can be estimated independently by exploiting the known direction of the laser beam and the determined position of the laser intersection with the scene $X_{\!L}$. As this point is seen on the image, the actual 3D point $X_{\!L}$ had to be in the line-of-sight of the camera and can therefore be deducted using a ray casting procedure. The location is computed by finding the first surface of the 3D model which is intersected by a ray originating in the camera center and passes through the location of the detected laser spot on the image. Subsequently, to obtain the location of the origin, the point $X_{\!L}$ expressed in camera frame is back-projected according to a known direction of the beam $v_{\!L}$ onto the plane $\mathcal{L}$ (Eqs.~\ref{eq:origin_laser_calib}). Once known, the scale can be determined by comparing the displacement $\hat{m}_{\!L} = \Vert \hat{O}_{\!L}\Vert$ with its \textit{a priori} known value $m_{\!L}$. 

\begin{equation}
\hat{O}_{\!L} = \boldsymbol{P}X_{\!L} - \frac{\boldsymbol{P}X_{\!L} \cdot c_z}{v_{\!L} \cdot c_z} v_{\!L}\:,
\label{eq:origin_laser_calib}
\end{equation}
where $c_z$ represents the optical axis of the camera.

\begin{figure}
	\centering
	\includegraphics[width=8cm]{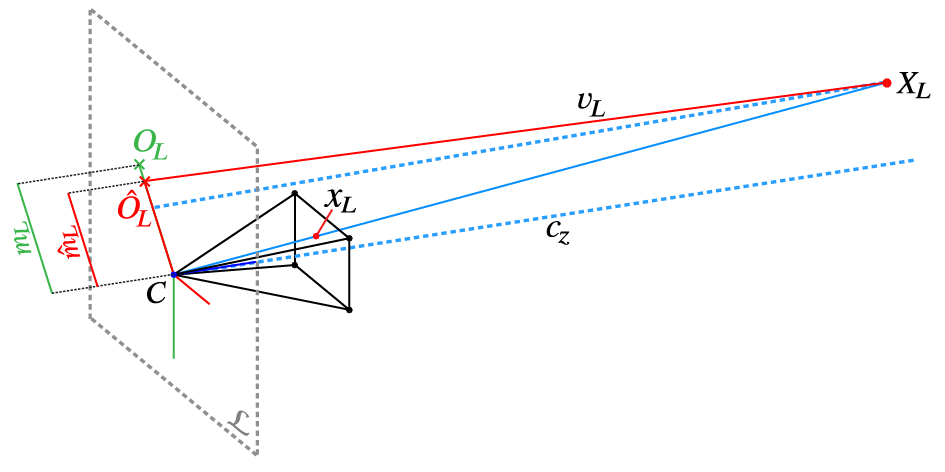}
	\caption{Scale estimation using the fully calibrated approach, based on the 3D model and optical image depicting the laser beam projection on the scene intersection with the scene.}
	\label{fig:measurement_calibrated}
\end{figure}

\begin{figure}[ht]
	\centering
	\includegraphics[width=8cm]{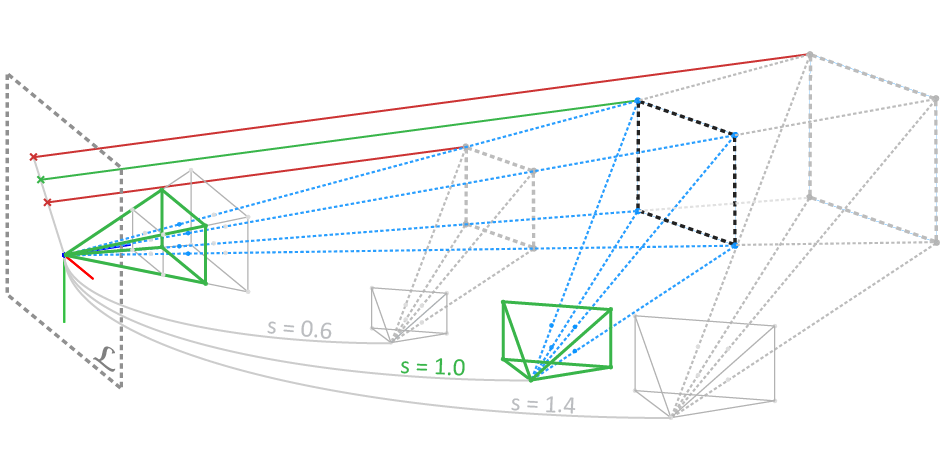}
	\caption{The effect of various scales affecting the 3D model (incorrect - light gray, correct - black) on the predicted location of the laser origin (incorrect - red, correct - green).}
	\label{fig:measurement_scale_ambiguity}
\end{figure}

Figure~\ref{fig:measurement_scale_ambiguity} depicts the effect of different model scales on the displacement of the predicted laser origin. Due to the scale ambiguity, all variations of the model (depict in light gray) are valid solutions of the 3D reconstruction process. As shown, the correct scale can be determined by comparing the displacement of a laser intersection point (blue) back-projected to the plane $\mathcal{L}$ with the \textit{a priori} known location of the laser origin. 

\subsubsection{Partially calibrated method}

While fully calibrated method enables an arbitrary laser setup, the required rigidity between the lasers and the camera can be extremely limiting in certain real scenarios. To alleviate this, we present an alternative approach, in which the required relation between the camera and the lasers is significantly reduced. The approach only requires two lasers to be parallel and equidistant to the camera. As opposed to the image scaling methods, the lasers do not have to be aligned with the optical axis of the camera. The scale of the model is therefore estimated by comparing a known perpendicular distance between the two parallel beams to the one estimated from the image and the model $\hat{d_{\!L}}$. To overcome the fact that the direction of the parallel beams wrt.~the camera is not known, we exploit the knowledge that the lasers are equidistant to the camera and approximate the direction with the direction of the vector connecting camera center and the middle point between the two points of lasers intersections with the model $X_{\!L_{\!1}}$ and $X_{\!L_{\!2}}$. As it is reasonable to expect for the depth discrepancy between the two points to be significantly smaller than the camera-scene distance, the approximation leads to a negligible error. Similar to the FCM, the location of laser intersections with the scene $X_{\!L_{\!1}}$ and $X_{\!L_{\!2}}$ are determined through a ray casting procedure and are affected by the same scale as the model and therefore affect the final estimated distance $\hat{d_{\!L}}$ by the same factor:

\begin{gather}
\cos{\alpha} = \frac{v_{\text{1,2}} \cdot v_{\text{CM}}}{\lvert v_{\text{1,2}} \rvert \lvert v_{\text{CM}} \rvert}\:,
\label{eq:partial_cos_alpha}
\\
\hat{d_{\!L}} = \sin{\alpha} \cdot \lvert v_{\text{1,2}} \rvert\:,
\label{eq:distance_laser_calib}
\end{gather}
where $v_{\!\text{1,2}}$ represents the vector between scene points $X_{\!L_{\!1}}$ and $X_{\!L_{\!2}}$ and $v_{\text{CM}}$ the vector connecting camera center with the middle point $X_{\!M}$.

\begin{figure}
	\centering
	\includegraphics[width=8cm]{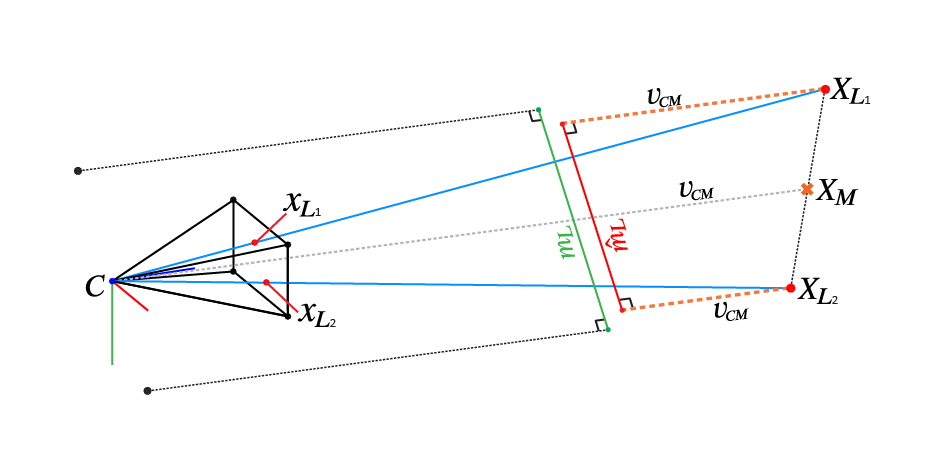}
	\caption{Scale estimation based on the 3D model and optical image of laser intersection with the scene using partially calibrated method.}
	\label{fig:measurement_partial}
\end{figure}

\section{Results}
\begin{figure}
	\centering
	\includegraphics[width=8cm]{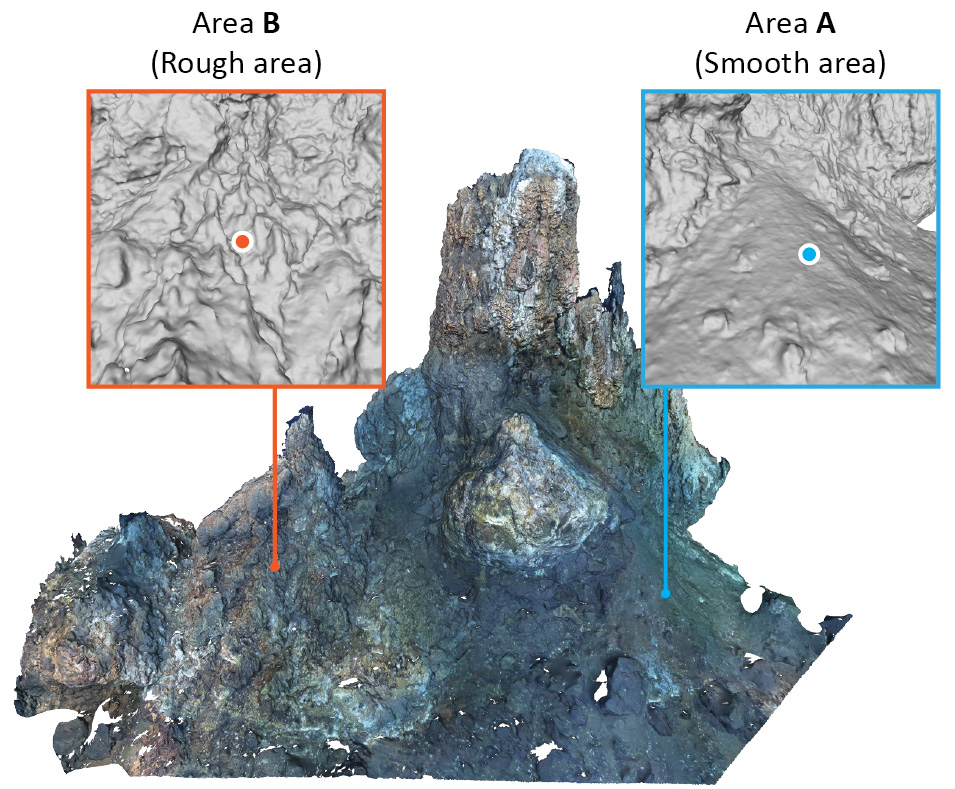}
	\caption{3D model of an underwater hydrothermal vent (Eiffel Tower at Lucky Strike vent field, Mid-Atlantic Ridge) used for model reconstruction evaluation at two marked areas. Data acquired during the 2015 MOMARSAT cruise (doi:10.17600/15000200).}
	\label{fig:results_area}
\end{figure}

To assess the applicability and accuracy of the two proposed approaches, partially and fully calibrated methods (PCM and FCM respectively), tests on both real and simulated scenario datasets were performed. To validate the performance using different laser configurations and acquisition conditions, we have used a real 3D model built using underwater imagery, as depicted in Fig.~\ref{fig:results_area}. Various laser measurements were generated as they would have been captured during an ROV survey. As the absolute scale of the model is not precisely known, for the purpose of this evaluation, it was assumed that the model and its scale are correct. Therefore, the performance can be evaluated by comparing the deviations of the estimated scales with the assumed (imposed) correct value of the scale of the model ($s\!=\!1$). This allowed us to confirm the correctness of our approaches, as well as analyze the effects various types and levels of noises have on the estimation. 

\begin{figure}
	\centering
	\includegraphics[width=8cm]{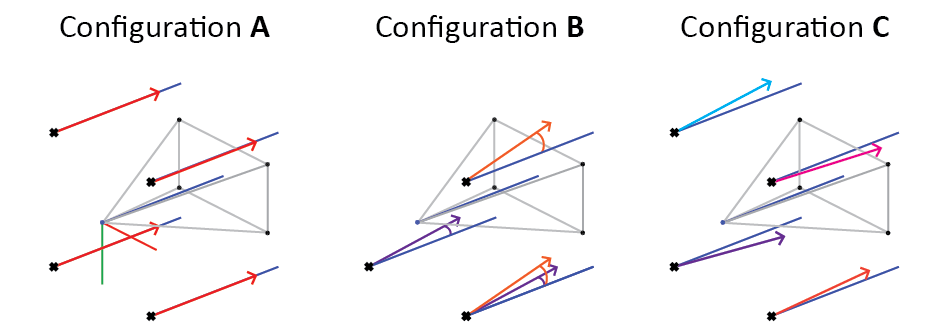}
	\caption{Various laser configurations used in evaluation: \textit{A}) Optical axis aligned laser beams; \textit{B}) Pair-wise parallel laser pairs; \textit{C}) Lasers with arbitrary origins and orientations. Blue lines represent the optical axis, and the remaining lines depict lasers which are parallel among themselves.}
	\label{fig:laser_configurations}
\end{figure}

Given our goal of developing methods usable in real scenarios, three separated laser configurations were devised (Fig.~\ref{fig:laser_configurations} to test the performance: 

\begin{enumerate}[A)]
    \item Lasers are parallel and aligned with the optical axis of the camera;
    \item Lasers are parallel and positioned equidistant from the camera center, but not aligned with the optical axis;
    \item Lasers have arbitrary positions and directions.
\end{enumerate}

To illustrate the advantages of our proposed methods in comparison to commonly used image-scaling approaches, the approach by Davis and Tusting~\citep{davis1991quantitative} was additionally evaluated, as one of the most versatile methods. The procedure requires four parallel lasers aligned with the optical axis of the camera as well as assumes scene flatness. By exploiting the known spacing between the laser spots on the image and displacement of laser origins from the optical center of the camera, distances between various points on the image can be computed for an arbitrary tilt and pan of the camera. As only laser configuration \textit{A} suffice the requirements of the method, and other configurations cause dramatic and unpredictable errors, we limit the reporting of the results for Davis approach to laser configuration \textit{A}. Another commonly used method presented by Pilgrim et al.~\citep{pilgrim2000rov} was not evaluated, as the method requires the restriction of the pose of the camera in either pan or tilt with respect to the scene, which can only be a reasonable restriction if the scene is flat (e.g., sea bottom), which is almost never the case in models reconstructed using \ac{SfM}.

\subsection{Data}
\begin{figure}
	\centering
	\includegraphics[height=4cm]{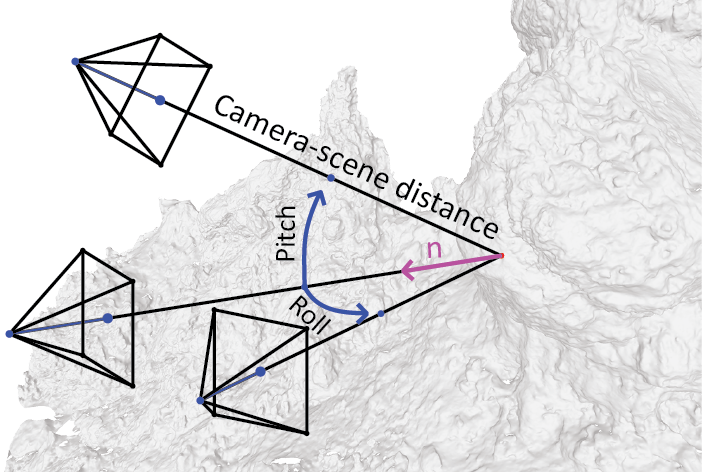}
	\caption{Definition of perspective angles and camera-scene distance used in the generation of the evaluation data.}
	\label{fig:camera_scene}
\end{figure}

The generation of image and laser data as they would have been recorded in real scenarios enabled us to simulate different perspective angles and camera-scene distances, and analyze their effects on the resulting estimations of scales. The real 3D model depicted in Fig.~\ref{fig:results_area} was used in this simulation. The 3D chimney was reconstructed from $908$ images of an underwater vent field at the deep-sea Lucky Strike area, collected during the MOMARSAT 2015 cruise (doi:10.17600/15000200). The model covers an area of approximately $\SI{200}{\meter\squared}$ with height range of $\sim\!\!\SI{13}{\meter}$. Assuming the 3D model has a correct scale, we can compute the location of laser spots and feature points as they would appear on the images taken from different poses and according to the pre-determined laser configurations. The number of feature points has been selected to reflect an average number of successfully matched features per image in underwater scenarios ($n\!=\!1500$). To mimic the various perspective angles of the camera, we generate views for which the image plane is not only perpendicular to the surface normal (at the point viewed by the principal point of the camera), but also at a wide range of angles. In total $289$ different views were created from different combinations of pitch and roll angles deviating from $\SI{-40}{\degree}$ to $\SI{40}{\degree}$ in $\SI{5}{\degree}$ steps (Fig.~\ref{fig:camera_scene}). If not specified differently, the camera-scene distance (i.e., distance between the camera center and the point of interest on the surface) has been kept constant at $\SI{3}{\meter}$; based on our experience, this is a reasonable assumption for typical ROV survey of the scene in this type of environments. 

The lasers have been positioned according to the configurations envisioned in different scenarios (Fig.~\ref{fig:laser_configurations}). In configuration \textit{A}, the lasers have been positioned at an equidistance of $\SI{10}{\centi\meter}$ from the camera center. For configuration \textit{B} two pairs of lasers, with a $\SI{10}{\centi\meter}$ perpendicular distance between the beams, have been used, positioned vertically and horizontally. The pairs are perfectly parallel but not aligned with the optical axis of the camera. Each of the pairs has been used independently to test the two most common scenarios, with laser scalers positioned either below or on the side of the camera. As both produced similar results we only present the results for the horizontal pair. 

Finally, the configuration \textit{C} reflects a real laser configuration used during the 2017 SUBSAINTES cruise (doi:10. 17600/17001000)~\citep{escartin2017}.
The laser set-up in the ROV VICTOR (IFREMER) used for image acquisition during this cruise was slightly misaligned, while the laser origins are placed at an approximately equal distance of $\SI{16.5}{\centi\meter}$ with slight rotation around the z-axis of the camera. 

\subsection{Terrain roughness}

\begin{figure}
	\centering
	\includegraphics[width=8cm]{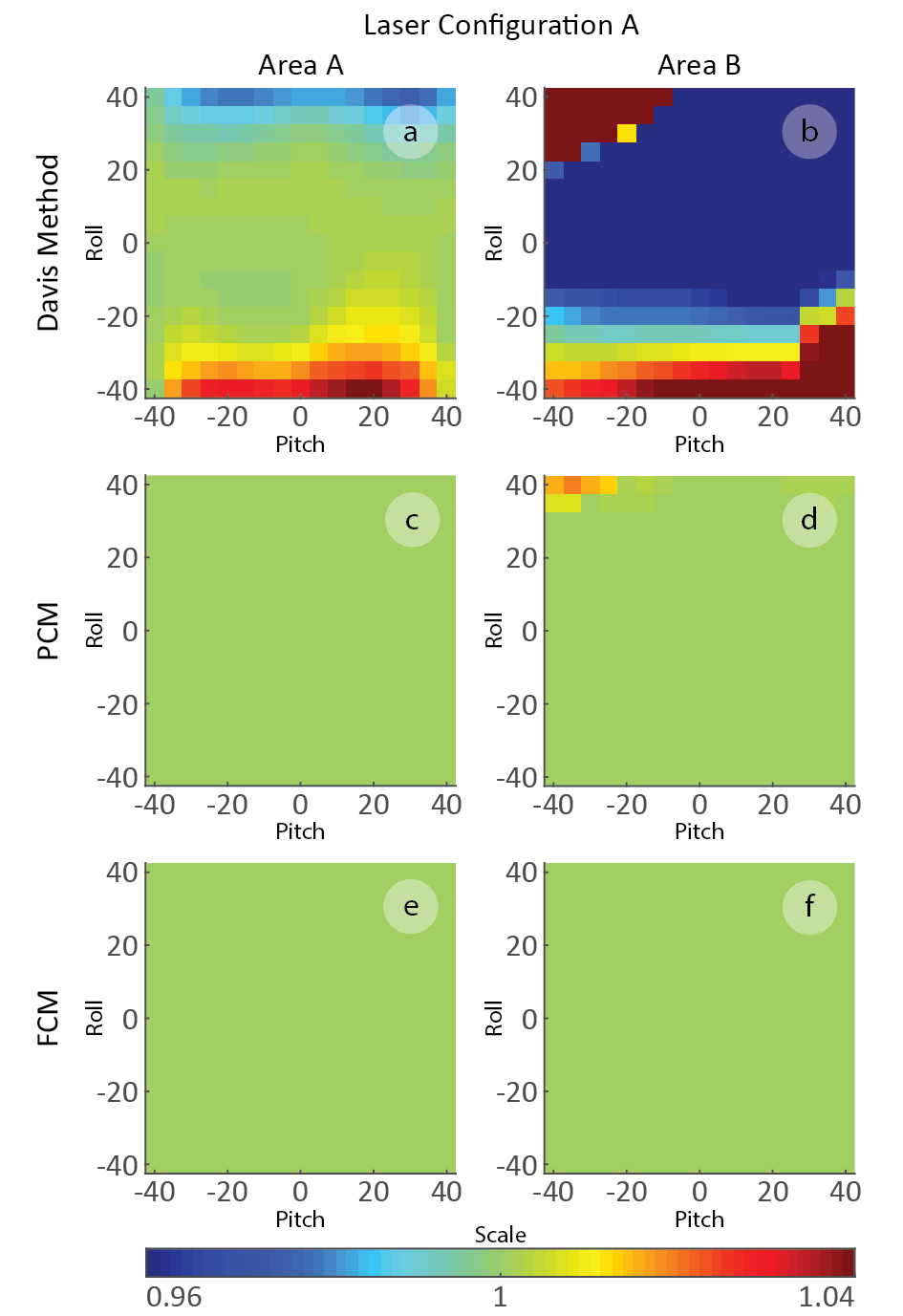}
	\caption{Estimated model scales at a smooth (area A) and rough area (area B) with various perspective angles and constant camera-scene distance ($d\!=\!\SI{3}{\meter}$) using Davis and Tusting~\citep{davis1991quantitative}, partially calibrated (PCM) and fully calibrated method (FCM). Lasers were aligned with the optical axis (configuration A).}
	\label{fig:results_roughness_AA}
\end{figure}

We first compare the results of estimated scales on two different types of terrain (smooth - \textit{Area A} and rough - \textit{Area B}) acquired from variety of perspective angles and laser configurations. Figure~\ref{fig:results_roughness_AA} presents the results obtained using laser configuration \textit{A} and with our two proposed methods as well as with the Davis approach. 

Comparing the errors among the methods, we notice that the Davis method is capable of estimating the correct scale only if the flatness assumption is only slightly violated, i.e. the area is nearly flat and the perspective angle is not too large (Fig.~\ref{fig:results_roughness_AA}a). As that is not the case on rough terrain (Fig.~\ref{fig:results_roughness_AA}b), the estimated scale varies significantly with different perspective angles, confirming the strong dependency of this method on scene geometry. On the other hand, our two methods correctly compensate for any changes in the viewing angle and terrain roughness. The laser direction approximation assumed in PCM does, however, cause a slight error - up to $1.5\%$ in extreme cases (e.g., rough terrain and large perspective angle - Fig.~\ref{fig:results_roughness_AA}d), situation in which the depth discrepancy between the two laser points is strongly boosted. Correctly estimated scale in all the cases, clearly shows the ability of the FCM to correctly compensate for the effects of terrain roughness and perspective angle (Figs.~\ref{fig:results_roughness_AA}e and~\ref{fig:results_roughness_AA}f). Additionally, it is important to re-emphasize, that image scaling methods require an additional association between the image points and the model in order to be able to estimate the scale. In our tests, we assumed perfect association, which is nearly impossible to achieve as it is a manual error-prone process. The actual results in real cases are therefore expected to be even worse.

In scenarios in which the lasers are not perfectly aligned with the camera (i.e., laser configurations \textit{B} and \textit{C}), the image scaling methods become unusable as the errors increase dramatically and unpredictably. For this reason, we only present the results of our proposed methods (PCM and FCM) for the remaining two configurations. Similarly, we limit the presented results to the rough terrain, as the methods will perform better (or equally) on flat areas. 

\begin{figure}
	\centering
	\includegraphics[width=8cm]{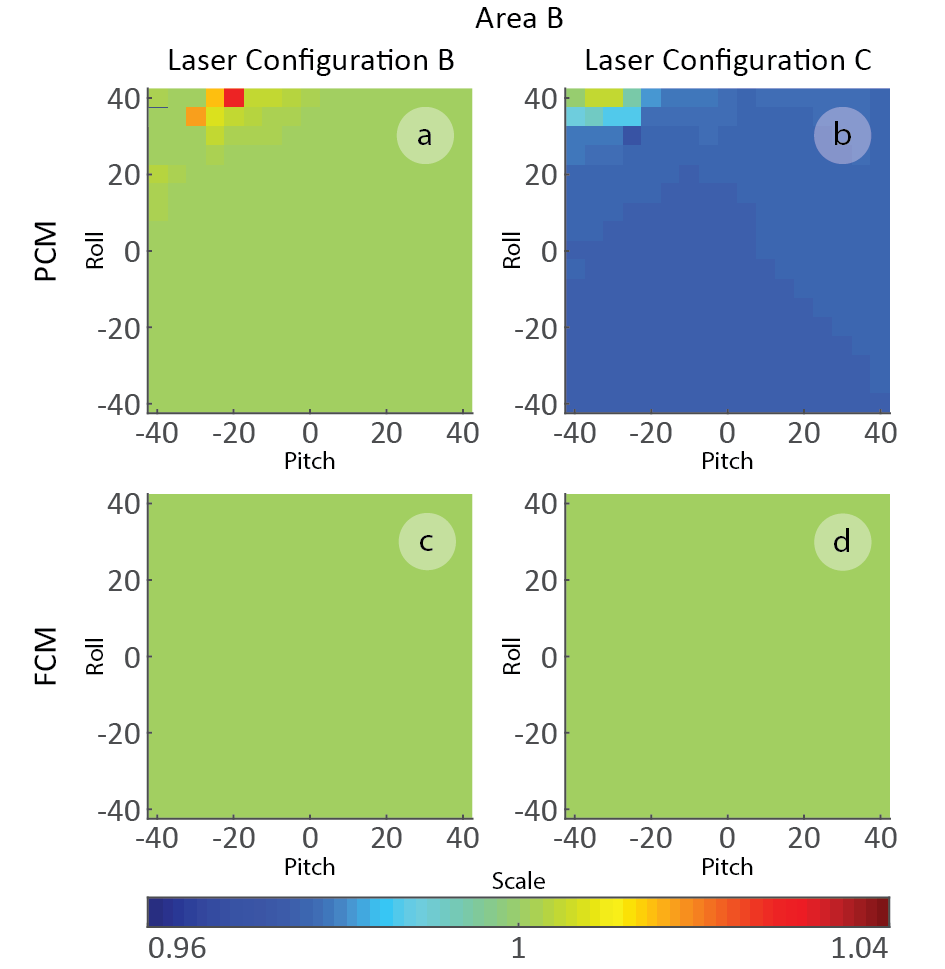}
	\caption{Estimated model scales at a rough area (area B) with various perspective angles and constant camera-scene distance ($d\!=\!\SI{3}{\meter}$) using partially calibrated (PCM) and fully calibrated method (FCM). Lasers were in configuration B and C.}
	\label{fig:results_roughness_PF}
\end{figure}

As seen in Figures~\ref{fig:results_roughness_PF}a and~\ref{fig:results_roughness_PF}c, both of our methods obtain good results with a laser configuration \textit{B}, in which the lasers are mounted parallel to each other. As in the previous cases, the partial method exhibits slight errors due to the assumed laser direction approximation. Analysis of data collected using laser configuration \textit{C}, the partial method fails, with results strongly affected by the irregularities in the parallelism. Instead, the full method (Fig.~\ref{fig:results_roughness_PF}d) correctly compensates these irregularities and yields correct results. 

\subsection{Laser direction approximation}
\begin{figure}
	\centering
	\includegraphics[width=8cm]{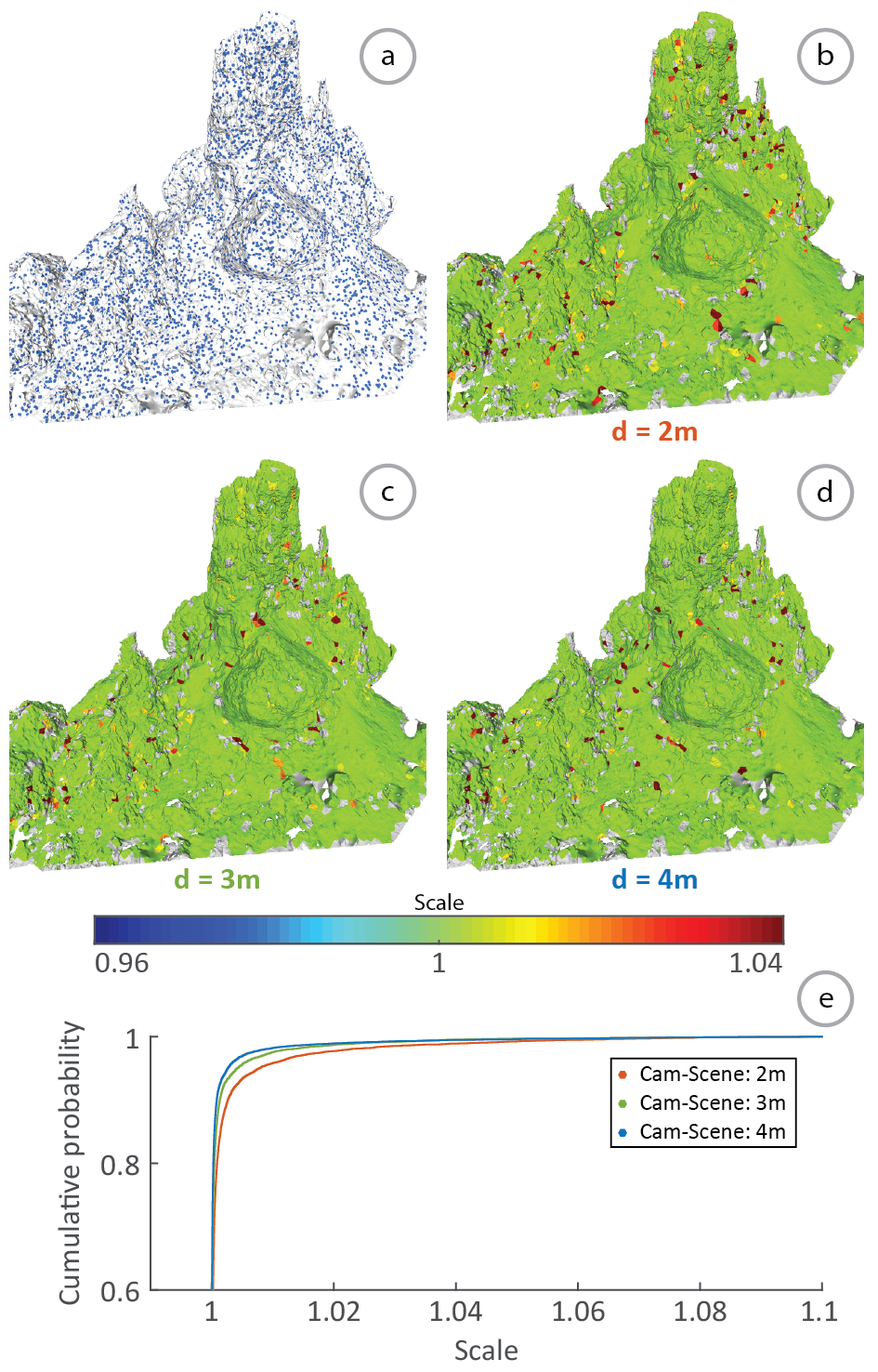}
	\caption{a) 10,000 random points used for estimating the scale across the model; b-d) Estimated model scales at various camera-scene distances with laser configuration B using a partially calibrated method; e) Cummulative probability distribution of estimated scales.}
	\label{fig:results_roughness_3D_P}
\end{figure}

To illustrate the influence of the depth difference between the two points hit by the laser beams and the camera-scene distance have on the result of the partial method, we have estimated the scale on $10,000$ randomly-selected points across the model (Fig.~\ref{fig:results_roughness_3D_P}a). For each point, the camera has been positioned at a  distance $d$ in the direction of the normal of the surface. Results obtained at three distances ($\SI{2}{\meter}$, $\SI{3}{\meter}$ and $\SI{4}{\meter}$), illustrated in Figs.~\ref{fig:results_roughness_3D_P}b-d, show that the error decreases with increasing distance of the camera (i.e., larger $d$).  This is especially visible in rougher areas, such as the top of the hydrothermal vent and the areas near previously mentioned area \textit{B}. As it is reasonable to assume that depth discrepancies between points in those areas will be bigger, the result indicates that the increased camera-scene distance decreases the effect depth discrepancies have on the accuracy of the results. We also document the cumulative distribution functions of these estimated scales obtained at different camera-scene distances (Fig.~\ref{fig:results_roughness_3D_P}e), from which it is noticeable that a higher percentage of points with scales closer to anticipated value of $1.0$ is obtained the further the camera is from the scene. 

\begin{figure}
	\centering
	\includegraphics[width=8cm]{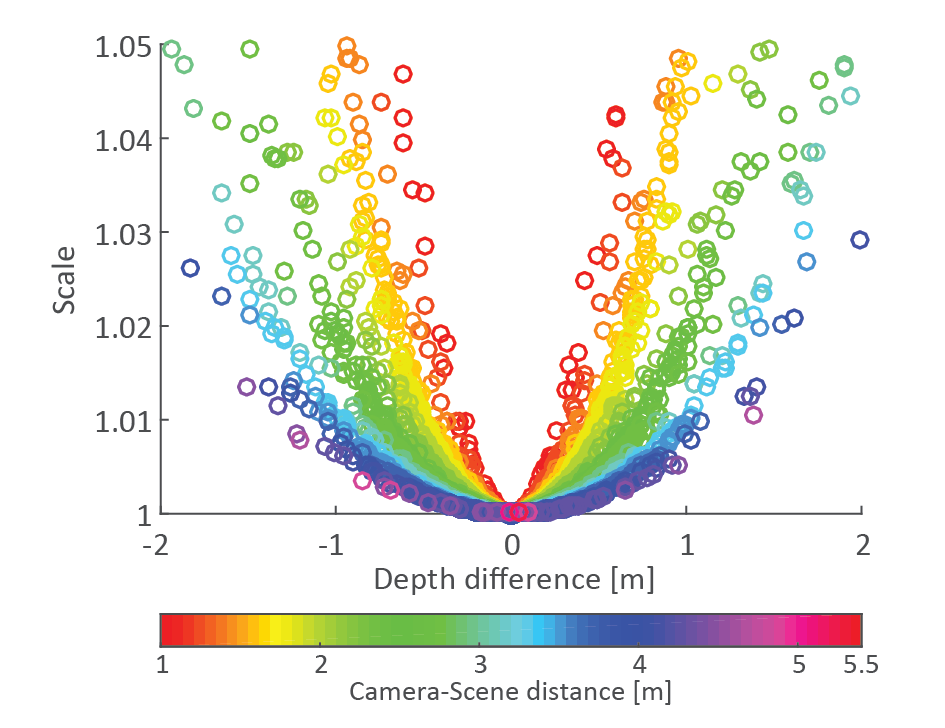}
	\caption{Estimated model scale (vertical axis) using a partially calibrated method at 10,000 random points, with varying camera-scene distances (color coded) and laser configuration B, as a function of the depth difference between the two points on the model (horizontal axis).}
	\label{fig:results_roughness_RND_D}
\end{figure}

The relation between the camera-scene distance and the depth difference can be clearly observed in Fig.~\ref{fig:results_roughness_RND_D}, which shows the estimated scale vs. depth difference, with color coded camera-scene distances. As expected, the error in the estimation grows with the increase in the depth discrepancies. Furthermore, we can see that the increase follows a parabola-shaped functions determined by the 
camera-scene distance. Short distances define a narrow parabola, and cause an increase in the error that is larger than that for longer distances. The shape and steepness of the parabolas is dependent on the displacement of the lasers from the camera origin, as well as their orientation with respect to the optical axis of the camera.

\subsection{Noise}

As collected data is never noise-free, we performed an additional analysis to evaluate the effects of the expected noise in feature and laser spot detection have on the scale estimation process. The experiment was performed on \textit{area B} of the model, with camera angles ranging from $\SI{-15}{\degree}$ to $\SI{15}{\degree}$ in pitch and roll; the range of view geometries which give consistent results in the ideal scenario (Fig.~\ref{fig:results_roughness_AA}). The observation distributions were modelled by assuming multivariate Gaussian distributions with dimension-independent noise for both feature and laser spot detections. For 2D features, the values were set matching those normally obtained in underwater scenarios ($\sigma_{f}\!=\!\big\{0.5\text{px}\text{,}\,1.0\text{px}\big\}$), while laser detection noise was defined by assuming $95\%$ accuracy of peak detection within one or two pixels ($\sigma_{l}\!=\!\big\{0.25\text{px}\text{,}\,0.5\text{px}\big\}$). As feature matches themselves are normally corrupted with a certain level of outliers, we have also performed experiments with various inlier/outlier ratios ($r\!=\!\big\{0\%\text{,}\,10\%\text{,}\,20\%\big\}$). Each of the tests has been repeated $500$ times.

\begin{figure}
	\centering
	\includegraphics[width=8cm]{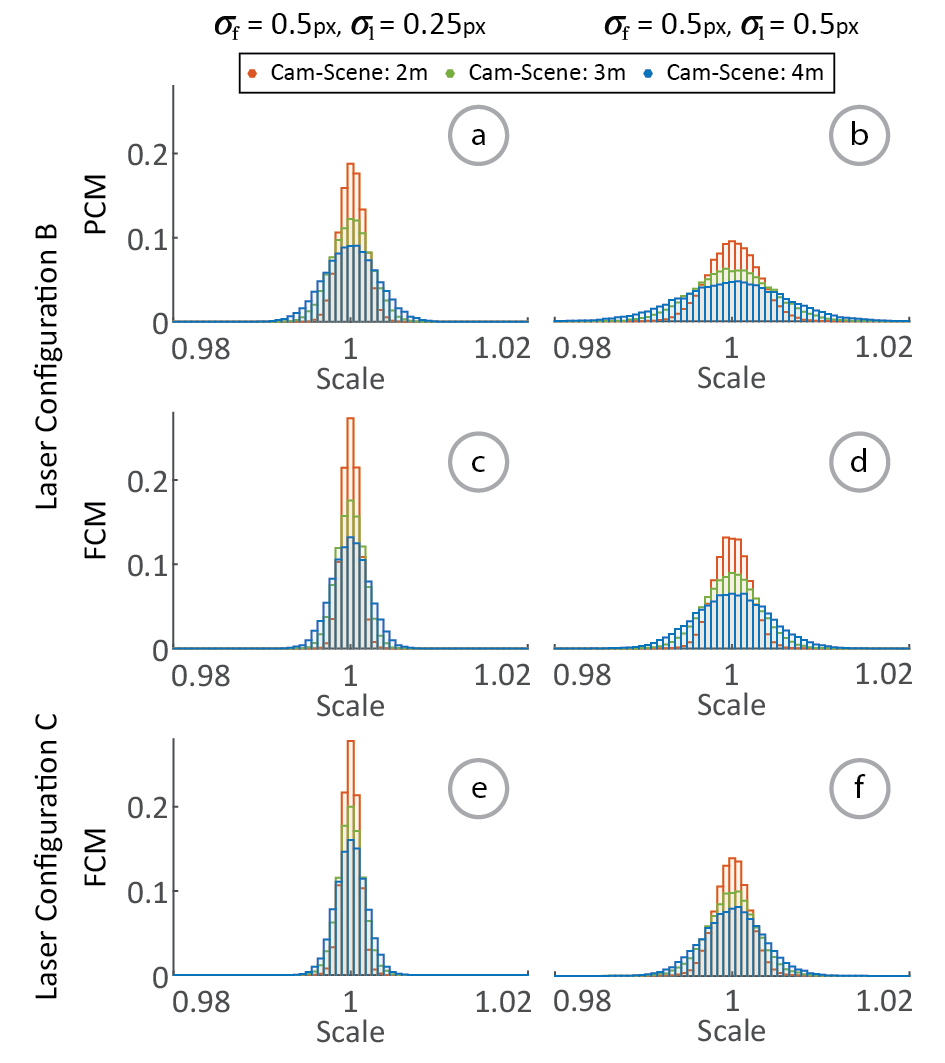}
	\caption{Distributions of estimated model scales with partially and fully calibrated methods at various noise levels induced into the location of detected features and laser spots. The results obtained at different camera-scene distances is depict with (2m - red; 3m - green; 4m - blue).}
	\label{fig:results_roughness_Noise}
\end{figure}

The resulting distributions of estimated scales with parallel and free laser configurations (i.e., configurations \textit{B} and \textit{C}) are presented in Table~\ref{table:results_roughness_Noise} with a subset of the results shown in Fig.~\ref{fig:results_roughness_Noise}. Given that the FCM requires only a single laser to obtain a scale estimate, results from separate lasers were fused by computing their average. The effect of such averaging can  be identified in Table~\ref{table:results_roughness_Noise}, where the results for a single laser (FCM - single) are shown side by side with the final averaged result (FCM - all).

\begin{table}
  \caption{The results obtained with various methods (PCM, FCM) with different levels of noise induced into the location of detected features and laser spots.}
  \scriptsize{\begin{tabular*}{\tblwidth}{@{} CCCCC@{} }
   \toprule
   Cam-Scene & \multicolumn{2}{c}{Configuration B} & \multicolumn{2}{c}{Configuration C}\\
   distance [m] & PCM & FCM - all & FCM - single & FCM - all\\
   \midrule
   \multicolumn{5}{l}{$\sigma_{f}\!=\!0.5$, $\sigma_{l}\!=\!0.25$}\\
   \midrule
   2 &
    $1.0\pm0.0014$ & $1.0\pm0.0010$ & $1.0\pm0.0019$ & $1.0\pm0.0010$\\
    3 &
    $1.0\pm0.0022$ & $1.0\pm0.0015$ & $1.0\pm0.0028$ & $1.0\pm0.0014$\\
    4 &
    $1.0\pm0.0030$ & $1.0\pm0.0021$ & $1.0\pm0.0034$ & $1.0\pm0.0017$\\
    
    \midrule
    \multicolumn{5}{l}{$\sigma_{f}\!=\!1.0$, $\sigma_{l}\!=\!0.25$}\\
    \midrule
    2 &
    $1.0\pm0.0014$ & $1.0\pm0.0010$ & $1.0\pm0.0019$ & $1.0\pm0.0010$\\
    3 &
    $1.0\pm0.0022$ & $1.0\pm0.0015$ & $1.0\pm0.0028$ & $1.0\pm0.0014$\\
    4 &
    $1.0\pm0.0030$ & $1.0\pm0.0021$ & $1.0\pm0.0034$ & $1.0\pm0.0017$\\
    \midrule
    \multicolumn{5}{l}{$\sigma_{f}\!=\!0.5$, $\sigma_{l}\!=\!0.5$}\\
    \midrule
    2 &
    $1.0\pm0.0028$ & $1.0\pm0.0020$ & $1.0\pm0.0038$ & $1.0\pm0.0020$\\
    3 &
    $1.0\pm0.0044$ & $1.0\pm0.0031$ & $1.0\pm0.0056$ & $1.0\pm0.0028$\\
    4 &
    $1.0\pm0.0059$ & $1.0\pm0.0042$ & $1.0\pm0.0069$ & $1.0\pm0.0034$\\
   \bottomrule
  \end{tabular*}}
	\label{table:results_roughness_Noise}
\end{table}

As expected, the uncertainty of estimated scales increases with the increasing noisiness of the laser detections, as each estimation is directly influenced by displacements in laser spot positions. Comparison of these results show that with noisy data the PCM method performs better than the FCM with a single laser point, but worse when multiple laser points are used instead. This occurs due to the averaging of independent scale estimates. As each laser produces a result that is independently affected by noise, the subsequent averaging reduces its effect. 

To some extent this can also be observed in the partial method with the simultaneous use of two laser points, which explains the improved results over the full method with the single laser. It is also clear that uncertainty of the scaling estimate also increases with the camera-scene distance, which is expected as errors on the image are  magnified when projected further from the camera.

In contrast, the noise corrupting the feature points used in the pose estimation, does not significantly affect the final scaling results. This is due to the use of \ac{BA} in the pose optimization, which is a maximum likelihood estimator mhen the image error is zero-mean and normally distributed, as it is the case in our tests. Similarly, the effects of outliers are mitigated by the use of a robust estimation method \ac{AC-RANSAC}~\citep{moisan2012automatic}. As the outliers do not follow a specific pattern, the iterative procedure successfully identifies and removes spurious matches, and hence the final estimate is unaffected. It is important to note that while the results obtained might indicate an extremely robust method to any discrepancy in the feature points, the approach is still vulnerable to a) outliers that obey the estimated geometric model, to b) the possibility of having a set of feature points which can be explained with multiple camera poses, or to both a) and b). However, this vulnerability can be reduced to a level that does not represent a practical concern, by ensuring that the set of features is well spread throughout the image.

\subsection{Real Scenario}

\begin{figure}
\centering
\includegraphics[width=8cm]{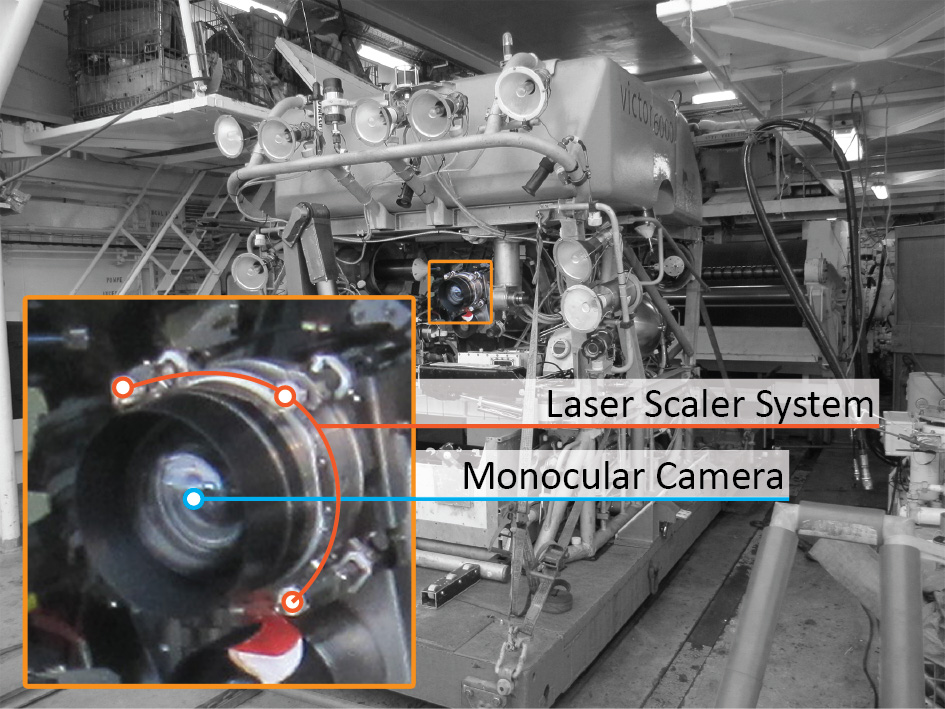}
\caption{ROV VICTOR 6000 (IFREMER) with enlarged camera and laser scaler system.}
\label{fig:res_victor_laser_cam}
\end{figure}

The fully calibrated method was used on a real dataset collected during the SUBSAINTES cruise (doi: 10.17600/ 17001000). Throughout the cruise, extensive seafloor imagery was collected using the \ac{ROV} VICTOR 6000 (IFREMER)~\citep{Michel2003} with a mounted monocular camera (Sony FCB-H11  with corrective optics and dome port), and a laser scaler with four laser beams positioned around the camera (Fig.~\ref{fig:res_victor_laser_cam}). The intrinsic parameters of the camera were determined using a standard calibration procedure~\citep{bouguet2008camera} assuming a pinhole model with the 3rd degree radial distortion model. Once calibrated, the camera parameters were kept constant through entire acquisition process. 

One of the main goals of this cruise is to identify, map, and measure indicators of displacement at the seafloor associated with a recent submarine earthquake~\citep{escartin2016} that occurred in the French Antilles, offshore Les Saintes Islands in 2004~\citep{feuillet2011mw}. These traces are visible in outcrops of an active submarine fault scarp at depths of up to $\sim\!\!\SI{1000}{\meter}$ below sea level, and that has been systematically mapped and surveyed. Imagery was used to obtain $\sim\!\!30$ three-dimentional models, that will be ultimately used to conduct measurements of displacement associated with the 2004 earthquake. Accurate and precise geological measurements thus require proper scaling. 


\begin{figure}
	\centering
	\begin{tabular}{@{}c}
		\includegraphics[width=8cm]{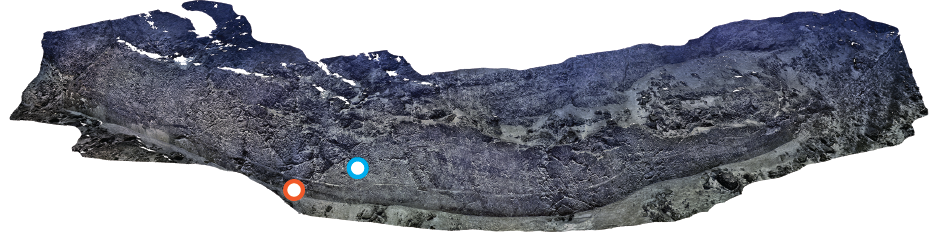}\\ 
		(a)\\
		\includegraphics[width=8cm]{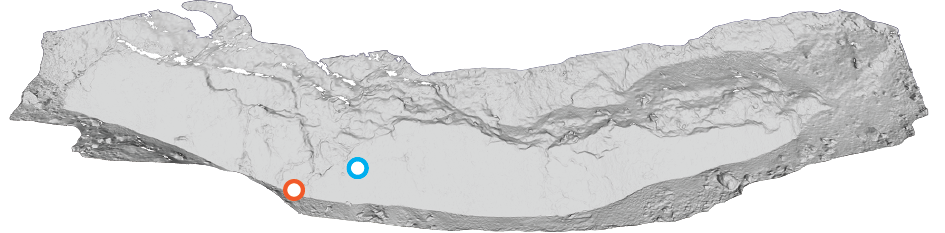}\\
		(b) 
	\end{tabular}
	\caption{a) Textured and b) triangle mesh representations of FPA 3D model, with marked areas of evaluation.}
	\label{fig:res_fpa_tex}
\end{figure}

The 3D models have been reconstructed using an adapted 3D reconstruction procedure consisting of multiple open-source solutions (OpenMVG~\citep{openMVG,moulon2013global}, OpenMVS~\citep{shen2013accurate,jancosek2014exploiting}, MVS-Texturing~\citep{waechter2014let}) as described in~\citep{hernandez2016autonomous}. Figure~\ref{fig:res_fpa_tex} depicts one such model, named FPA, which has been reconstructed from a total of $218$ images with the resolution of $1920\times 1080$. This particular outcrop was already imaged during a prior cruise (ODEMAR, doi:10.17600/13030070)~\citep{escartin2016}.

\begin{figure}
\centering
\includegraphics[width=8cm]{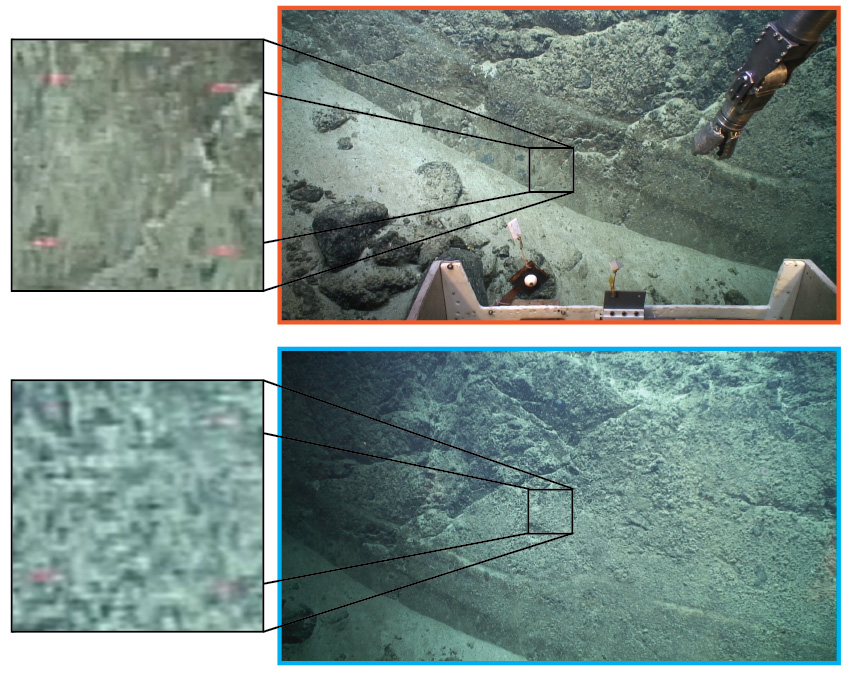}
\caption{Example of images from the two areas of evaluation with visible laser projections on the scene.}
\label{fig:optical_w_lasers}
\end{figure}

As the FPA model was reconstructed only using optical images acquired by a monocular camera, the scale of the resulting model is ambiguous, i.e.,~estimated parameters can be multiplied with an arbitrary factor and still produce equal projections of the model on the images~\citep{lourakis2013accurate,zisserman2004multiple}. In order to obtain a proper scale of the model, images containing laser beams projected on the surface of the scene can be used through one of our proposed methods. During the SUBSAINTES cruise, such images have been collected in addition to the ones already used in the reconstruction process. Six images with clearly noticeable laser spots (Fig.~\ref{fig:optical_w_lasers}) have been selected from the center of the 3D model, at two different locations as indicated in  Fig.~\ref{fig:optical_w_lasers}. The images were collected at camera-scene distances of approximately $\SI{3}{\meter}$ and $\SI{4}{\meter}$ respectively while keeping the camera intrinsic parameters constant and equal to the ones used in the acquisition process. Subsequently, the laser spots locations have been marked manually (with the guidance of simple color thresholding) with expected error to be on average between $1\:\text{px}$ and $2\:\text{px}$. Due to multiple changes in the vehicle payload throughout the cruise, the lasers became misaligned and therefore a fully calibrated method was used to obtain the scale of the model. 

\begin{figure}
\centering
\includegraphics[width=8cm]{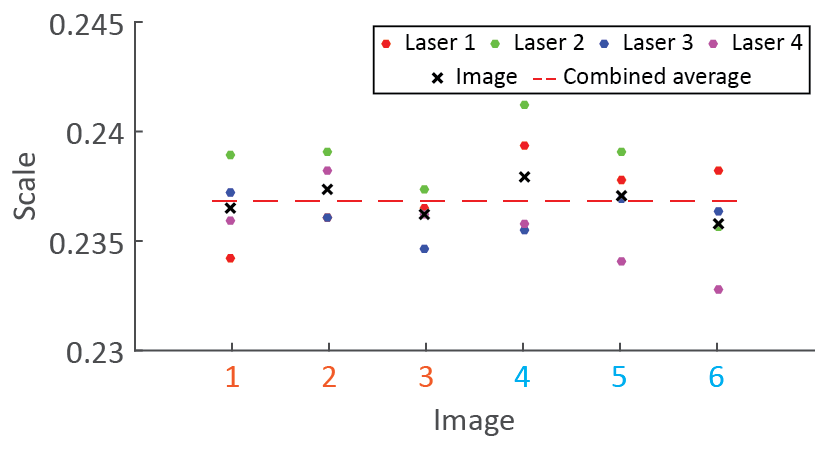}
\caption{Estimated scales for FPA model, per laser and per image, using fully calibrated method. Colour of image numbers (x axis) corresponds to locations shown in Fig.\ref{fig:optical_w_lasers}.}
\label{fig:results_fpa_lasers}
\end{figure}

Given that the setup consisted of four lasers, the FCM method computed four independent estimates of the model's scale per image. As we have shown in the previous experiments, averaging these independent results further reduces the effects of errors in the detection processes, leading to a better constrained final solution. The scaling results for each of the $6$ selected images are presented in Table~\ref{table:results_fpa_lasers} and Fig.\ref{fig:results_fpa_lasers}. In this figure, the scale estimates obtained for each laser beam are depict as circles, while the final estimate per image is marked with a black cross (x). The average of all the values obtained is additionally shown by a red dashed line. 

The average value of the scale of the FPA model estimated per image was $0.237 \pm 0.001$ which represents $0.3\%$ of the scale value. The obtained result implies that each unit in the current model is equal to $\SI{0.237}{\meter}$ or alternatively, the model has to be scaled with a factor $4.22$ to obtain a metric result. Comparing the deviations of scale estimates for image sets $1$-$3$ and $4$-$6$, the correlation between increasing camera-scene distance and increased uncertainty is apparent and consistent with previous result from generated data.  

\begin{table}
  \caption{Estimated FPA model's scale using fully calibrated method and simplistic direct 3D approach. Reported numbers represent the ratio between the model's unit and a meter - each measurement has to be multiplied with the inverse of the ratio to obtain metric result.}
  \scriptsize{\begin{tabular*}{\tblwidth}{@{} CCCCCCCC@{} }
   \toprule
    & Cam-Scene & \multicolumn{4}{c}{FCM (per laser)} & FCM & Direct 3D\\
    & distance [m] & $L_{\:1}$ & $L_{\:2}$ & $L_{\:3}$ & $L_{\:4}$ & (all) & (all)\\
   \midrule
1 & $3.05$ & $0.234$ & $0.239$ & $0.237$ & $0.236$ & $0.237\pm0.002$ & $0.235\pm0.009$ \\
2 & $3.06$ & $0.236$ & $0.239$ & $0.236$ & $0.238$ & $0.237\pm0.002$ & $0.236\pm0.008$ \\
3 & $3.05$ & $0.237$ & $0.237$ & $0.235$ & $0.236$ & $0.236\pm0.001$ & $0.235\pm0.008$ \\
4 & $3.90$ & $0.239$ & $0.241$ & $0.236$ & $0.236$ & $0.238\pm0.003$ & $0.236\pm0.013$ \\
5 & $3.91$ & $0.238$ & $0.239$ & $0.237$ & $0.234$ & $0.237\pm0.002$ & $0.236\pm0.013$ \\
6 & $3.60$ & $0.238$ & $0.236$ & $0.236$ & $0.233$ & $0.236\pm0.002$ & $0.234\pm0.010$ \\
   \bottomrule
  \end{tabular*}}
	\label{table:results_fpa_lasers}
\end{table}

\begin{figure}
\centering
\includegraphics[width=8cm]{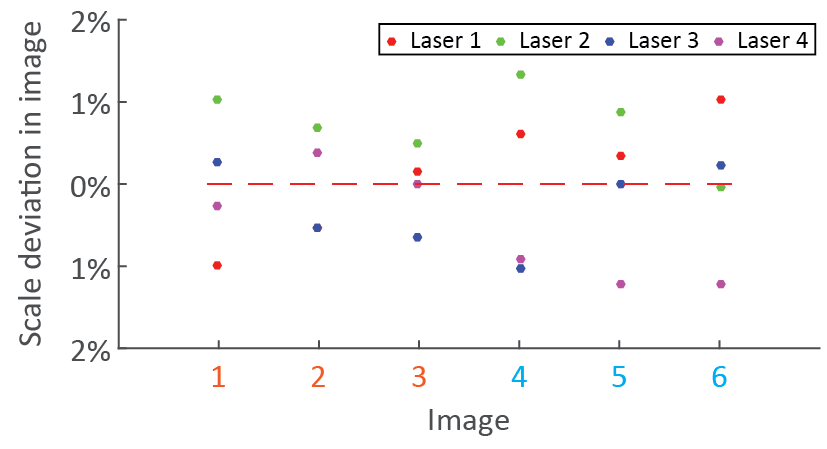}
\caption{Deviation of estimated FPA model's scales, using our fully calibrated method FCM, and for each laser in each image.}
\label{fig:results_fpa_lasers_each}
\end{figure}

The analyses of scaling deviations computed for each laser with respect to the final estimated scale per image (Fig.~\ref{fig:results_fpa_lasers_each}) shows that independent evaluations deviate about $0.6\%$ with a maximum deviation of $1.3\%$ for laser $2$ in image $4$. These results are again in agreement  with the results previously computed with the validation data on the hydrothermal vent in Fig.\ref{fig:results_area}. 

To further show the robustness and usefulness of our approach, we compare our results to the ones that would have been obtained if our method was not available. As the non-alignment of lasers with the optical axis would have prevented the use of both image-scaling methods (Pilgrim et al.~\citep{pilgrim2000rov} and Davis and Tusting~\citep{davis1991quantitative}), the only option available would have been a manual and somewhat simplistic approach still widely used in laser photogrammetry~\citep{kocak2004remote,rowe2008laser,robert2017new, pilgrim2000rov}. This involves manual identification of laser intersection points with the scene on the 3D model, and assuming pair-wise Euclidean distances to be the actual distances between the laser pairs. In order to compare our results with the best possible outcome of this simplistic approach, we determined the points on the model using a ray-casting technique, effectively completely eliminating the extremely error-prone human step. The results averaged over $4$ laser pairs are presented in the last column of Table~\ref{table:results_fpa_lasers} (Direct 3D). We can see that the results of different laser pairs are much more incoherent ($4.3\%$ deviation compared to $0.6\%$ in the case of fully calibrated method). We also note that the results of such simplistic method are extremely dependent on the perspective angle of the camera, the degree of misalignment of the lasers, as well as errors induced by manual point selection. As shown with the validation tests, our fully-calibrated method remains unaffected.

\section{Conclusions}

This paper introduced two novel methods for automatic scaling of SfM-based 3D reconstructions using laser scalers, that are applicable for routine underwater surveys with ROVs or AUVs. Both methods were validated using a series of generated datasets based on an underwater 3D model derived from submarine field imagery, and showed its applicability in real scenario using a dataset collected during a recent cruise (SUBSAINTES 2017). 

The two approaches presented here, namely fully and partially calibrated method, overcome a multitude of restrictions imposed by prior laser photogrammetry methods (e.g., laser alignment with the optical axis of the camera, perpendicularity of lasers with the scene). These methods, within the step of pose estimation, also remove the need for manual identification of identical points on the image and 3D model, an extremely time-consuming and error-prone processing step.

Each of the two methods is designed to address the different type of laser setup, encompassing the variety of most commonly used setups in real underwater scenarios. The fully calibrated method is applicable to arbitrary laser setups, with known geometric relations between the camera and the lasers. The ability to compensate for any misalignments enables accurate scaling in a wider variety of circumstances, such as the manipulation of equipment between surveys during a cruise and precluding strict parallelism. We thus propose a partially-calibrated method, which significantly reduces the camera-laser rigidity constraints, that may be otherwise too restrictive in real scenarios. This approach requires parallel lasers but alleviates the need for a time-consuming calibration process. The partially-calibrated method can thus be used to accurately and automatically scale 3D models built with data acquired using ROVs, including smaller shallow-water ones. Nowadays readily available pre-calibrated underwater laser scalers need only to be placed near the optical camera. 

To robustly validate the performance of the methods, a real 3D model of an underwater hydrodynamic vent was used to generate laser and image information as it would have been obtained from various laser configurations, camera viewing angles and camera-scene distances. We tested our methods with three laser configurations (i.e., aligned with the optical axis of the camera, parallel but misaligned with the optical axis and freely oriented) which can account for nearly all possible laser setups in real seafloor surveying situations using ROVs and AUVs. The initial evaluation was performed on two different types of terrain (smooth and rough), and demonstrated the advantages provided by the two proposed approaches relative to previously used image-scaling methods. Our methods can be used in the field, with misaligned or freely oriented lasers, and with extreme camera angles during image acquisitions, reaching up to $\SI{40}{\degree}$ in both pitch and roll.

While the fully calibrated method yielded robust results under all tested circumstances, the partially calibrated method was affected by a slight error ($2.9\%$ in the most extreme case) due to the approximation used for determining the laser direction. We further analyzed the effect of the approximation by evaluating $10,000$ randomly selected points. We demonstrate that scaling errors depend on the depth difference between the two points of laser-scene intersection, and that this effect decreases with an increasing camera-scene distance. The consequences of inevitable noise in feature and laser spot detection uncertainty were also examined, together with the effects of potential errors in feature matching (outliers). Due to the specificity of the algorithms used, the noise and potential outliers in the feature detection and matching process did not have a significant effect on the results, while the noise induced on the position of laser spots did directly influence the estimations. As expected, increases in camera-scene distance results in higher errors in the estimation, as the displacements are magnified with distance. Additionally we compared the results obtained from a single laser measurement with the average obtained from all and demonstrated that such fusion further reduces the effects of noise.

Finally we report on the application of the fully-calibrated method to determine the scale of a model built using images from a geologic outcrop, recorded during the SUBSAINTES cruise. Six images with clearly visible laser spots have been selected from two different model locations, and used to independently determine the scale of the model. The average scale estimated using our fully calibrated method was $0.237$ with the standard deviation of $0.3\%$ between the results from various images. The average deviation of estimated scales by independent lasers was $0.6\%$ with the maximum deviation of $1.3\%$. We also documented that images acquired at a longer camera-scene distance exhibited in bigger deviations of estimated scales, as predicted from the validation test results.

The results of our two methods were also compared to those that would have been obtained without the availability of our method. Due to laser non-alignment with the optical axis of the camera, the only approach possible would be a somewhat simplistic method which involves manual identification of laser intersection points with the 3D model, and assumes that the pair-wise Euclidean distances are the actual distances between the laser pairs. To predict the best possible outcome, we automatically determined these correspondences, alleviating any additionally induced errors. The results from the simplistic scale method show a much more important deviation than that of our method ($4.3\%$ vs. $0.6\%$, respectively). Based on our results we also stress that the results of such simplistic methods are extremely dependent on the perspective angle of the camera and the degree of misalignment of the lasers, which is not the case for our fully-calibrated method. Finally, these methods can be used universally as they are based on standard sensors available for ROVs and AUVs (cameras and laser scalers), do not require any dedicated hardware, and can be applied to legacy data.

Although the presented methods are designed to be independent of the laser spot detection approach used, we showed that its performance directly influences the scale estimation accuracy. In the reported results, we identified the location of the spots manually albeit with help of simple color thresholding. While relatively accurate, this manual process is time consuming. An effort is currently ongoing on automatizing the detection of the laser spots, which will facilitate the ability to perform scale estimation on larger number of images.

\section*{Acknowledgement}
This study is based on results from the MOMARSAT 2015 and SUBSAINTES 2017 cruises, that deployed the ROV VICTOR 6000 (IFREMER, France) for image acquisition used here. These cruises (ship and ROV time) were funded by the French Ministry of Research. We commend the work of the crew, officers, and engineers that participated on these cruises and made possible this data acquisition. Partial funding was provided 
by the European Union's Horizon 2020 project ROBUST (grant agreement  690416-H2020-CS5-2015-onestage) (to K.~Isteni\v{c}), project Eurofleets Plus (grant agreement 824077), the Spanish Ministry of Education, Culture and Sport under project CTM2017-83075-R (to R. Garcia and N. Gracias), the ANR SERSURF Project (ANR-17-CE31-0020, France) (to J.~Escart\'in and A.~Arnaubec), and the Institut de Physique du Globe de Paris (to J.~Escart\'in).

\printcredits

\bibliographystyle{model2-names}

\bibliography{biblio}

\begin{thebibliography}{58}
\expandafter\ifx\csname natexlab\endcsname\relax\def\natexlab#1{#1}\fi
\providecommand{\url}[1]{\texttt{#1}}
\providecommand{\href}[2]{#2}
\providecommand{\path}[1]{#1}
\providecommand{\DOIprefix}{doi:}
\providecommand{\ArXivprefix}{arXiv:}
\providecommand{\URLprefix}{URL: }
\providecommand{\Pubmedprefix}{pmid:}
\providecommand{\doi}[1]{\href{http://dx.doi.org/#1}{\path{#1}}}
\providecommand{\Pubmed}[1]{\href{pmid:#1}{\path{#1}}}
\providecommand{\bibinfo}[2]{#2}
\ifx\xfnm\relax \def\xfnm[#1]{\unskip,\space#1}\fi
\bibitem[{Agarwal et~al.(2009)Agarwal, Snavely, Simon, Seitz and
  Szeliski}]{agarwal2009building}
\bibinfo{author}{Agarwal, S.}, \bibinfo{author}{Snavely, N.},
  \bibinfo{author}{Simon, I.}, \bibinfo{author}{Seitz, S.M.},
  \bibinfo{author}{Szeliski, R.}, \bibinfo{year}{2009}.
\newblock \bibinfo{title}{Building rome in a day}, in: \bibinfo{booktitle}{2009
  IEEE 12th International Conference on Computer Vision}, pp.
  \bibinfo{pages}{72--79}.
\newblock \DOIprefix\doi{10.1109/ICCV.2009.5459148}.
\bibitem[{Anderson and Gaston(2013)}]{anderson2013light}
\bibinfo{author}{Anderson, K.}, \bibinfo{author}{Gaston, K.J.},
  \bibinfo{year}{2013}.
\newblock \bibinfo{title}{Lightweight unmanned aerial vehicles will
  revolutionize spatial ecology}.
\newblock \bibinfo{journal}{Frontiers in Ecology and the Environment}
  \bibinfo{volume}{11}, \bibinfo{pages}{138--146}.
\newblock \DOIprefix\doi{10.1890/120150}.
\bibitem[{Bergmann et~al.(2011)Bergmann, Langwald, Ontrup, Soltwedel, Schewe,
  Klages and Nattkemper}]{bergmann2011mega}
\bibinfo{author}{Bergmann, M.}, \bibinfo{author}{Langwald, N.},
  \bibinfo{author}{Ontrup, J.}, \bibinfo{author}{Soltwedel, T.},
  \bibinfo{author}{Schewe, I.}, \bibinfo{author}{Klages, M.},
  \bibinfo{author}{Nattkemper, T.W.}, \bibinfo{year}{2011}.
\newblock \bibinfo{title}{Megafaunal assemblages from two shelf stations west
  of svalbard}.
\newblock \bibinfo{journal}{Marine Biology Research} \bibinfo{volume}{7},
  \bibinfo{pages}{525--539}.
\newblock \DOIprefix\doi{10.1080/17451000.2010.535834}.
\bibitem[{Bingham et~al.(2010)Bingham, Foley, Singh, Camilli, Delaporta,
  Eustice, Mallios, Mindell, Roman and Sakellariou}]{bingham2010robotic}
\bibinfo{author}{Bingham, B.}, \bibinfo{author}{Foley, B.},
  \bibinfo{author}{Singh, H.}, \bibinfo{author}{Camilli, R.},
  \bibinfo{author}{Delaporta, K.}, \bibinfo{author}{Eustice, R.},
  \bibinfo{author}{Mallios, A.}, \bibinfo{author}{Mindell, D.},
  \bibinfo{author}{Roman, C.}, \bibinfo{author}{Sakellariou, D.},
  \bibinfo{year}{2010}.
\newblock \bibinfo{title}{Robotic tools for deep water archaeology: Surveying
  an ancient shipwreck with an autonomous underwater vehicle}.
\newblock \bibinfo{journal}{Journal of Field Robotics} \bibinfo{volume}{27},
  \bibinfo{pages}{702--717}.
\newblock \DOIprefix\doi{10.1002/rob.20350}.
\bibitem[{Bodenmann et~al.(2017)Bodenmann, Thornton and
  Ura}]{bodenmann2017generation}
\bibinfo{author}{Bodenmann, A.}, \bibinfo{author}{Thornton, B.},
  \bibinfo{author}{Ura, T.}, \bibinfo{year}{2017}.
\newblock \bibinfo{title}{Generation of high-resolution three-dimensional
  reconstructions of the seafloor in color using a single camera and structured
  light}.
\newblock \bibinfo{journal}{Journal of Field Robotics} \bibinfo{volume}{34},
  \bibinfo{pages}{833--851}.
\newblock \DOIprefix\doi{10.1002/rob.21682}.
\bibitem[{Bouguet(2008)}]{bouguet2008camera}
\bibinfo{author}{Bouguet, J.Y.}, \bibinfo{year}{2008}.
\newblock \bibinfo{title}{Camera calibration toolbox for matlab (2008)}.
\newblock \bibinfo{journal}{URL http://www. vision. caltech.
  edu/bouguetj/calib\_doc} \bibinfo{volume}{1080}.
\bibitem[{Caimi and Tusting(1987)}]{caimi1987application}
\bibinfo{author}{Caimi, F.M.}, \bibinfo{author}{Tusting, R.F.},
  \bibinfo{year}{1987}.
\newblock \bibinfo{title}{Application of lasers to ocean research and image
  recording systems}, in: \bibinfo{booktitle}{Proceedings of the International
  Conference on LASERS}, \bibinfo{organization}{STS Press McLean, Virginia}.
  pp. \bibinfo{pages}{518--524}.
\bibitem[{Caimi et~al.(1993)Caimi, Blatt, Grossman, Smith, Hooker, Kocak and
  Gonzalez}]{caimi1993advanced}
\bibinfo{author}{Caimi, M.}, \bibinfo{author}{Blatt, J.H.},
  \bibinfo{author}{Grossman, B.G.}, \bibinfo{author}{Smith, D.},
  \bibinfo{author}{Hooker, J.}, \bibinfo{author}{Kocak, D.M.},
  \bibinfo{author}{Gonzalez, F.}, \bibinfo{year}{1993}.
\newblock \bibinfo{title}{Advanced underwater laser systems for ranging, size
  estimations, and profiling}.
\newblock \bibinfo{journal}{Marine Technology Society Journal}
  \bibinfo{volume}{27}, \bibinfo{pages}{31--41}.
\bibitem[{Campos et~al.(2016)Campos, Gracias and Ridao}]{campos2016underwater}
\bibinfo{author}{Campos, R.}, \bibinfo{author}{Gracias, N.},
  \bibinfo{author}{Ridao, P.}, \bibinfo{year}{2016}.
\newblock \bibinfo{title}{Underwater multi-vehicle trajectory alignment and
  mapping using acoustic and optical constraints}.
\newblock \bibinfo{journal}{Sensors} \bibinfo{volume}{16},
  \bibinfo{pages}{387}.
\newblock \DOIprefix\doi{10.3390/s16030387}.
\bibitem[{Cocito et~al.(2003)Cocito, Sgorbini, Peirano and
  Valle}]{cocito2003bio}
\bibinfo{author}{Cocito, S.}, \bibinfo{author}{Sgorbini, S.},
  \bibinfo{author}{Peirano, A.}, \bibinfo{author}{Valle, M.},
  \bibinfo{year}{2003}.
\newblock \bibinfo{title}{3-d reconstruction of biological objects using
  underwater video technique and image processing}.
\newblock \bibinfo{journal}{Journal of Experimental Marine Biology and Ecology}
  \bibinfo{volume}{297}, \bibinfo{pages}{57--70}.
\newblock \DOIprefix\doi{10.1016/S0022-0981(03)00369-1}.
\bibitem[{Davis and Tusting(1991)}]{davis1991quantitative}
\bibinfo{author}{Davis, D.}, \bibinfo{author}{Tusting, R.},
  \bibinfo{year}{1991}.
\newblock \bibinfo{title}{Quantitative benthic photography using laser
  calibrations}.
\newblock \bibinfo{journal}{Undersea World, San Diego, California} .
\bibitem[{E.~Rowe and Dawson(2008)}]{rowe2008laser}
\bibinfo{author}{E.~Rowe, L.}, \bibinfo{author}{Dawson, S.},
  \bibinfo{year}{2008}.
\newblock \bibinfo{title}{Laser photogrammetry to determine dorsal fin size in
  a population of bottlenose dolphins from doubtful sound, new zealand}.
\newblock \bibinfo{journal}{Australian Journal of Zoology}
  \bibinfo{volume}{56}, \bibinfo{pages}{239--248}.
\newblock \DOIprefix\doi{10.1071/ZO08051}.
\bibitem[{Eltner and Schneider(2015)}]{eltner2015analysis}
\bibinfo{author}{Eltner, A.}, \bibinfo{author}{Schneider, D.},
  \bibinfo{year}{2015}.
\newblock \bibinfo{title}{Analysis of different methods for 3d reconstruction
  of natural surfaces from parallel-axes uav images}.
\newblock \bibinfo{journal}{The Photogrammetric Record} \bibinfo{volume}{30},
  \bibinfo{pages}{279--299}.
\newblock \DOIprefix\doi{10.1111/phor.12115}.
\bibitem[{Escart\'in et~al.(2017)Escart\'in, Le~Friant and
  Feuillet}]{escartin2017}
\bibinfo{author}{Escart\'in, J.}, \bibinfo{author}{Le~Friant, A.},
  \bibinfo{author}{Feuillet, N.}, \bibinfo{year}{2017}.
\newblock \bibinfo{title}{Subsaintes cruise report, n/o l'atalante - rov victor
  - auv asterx}.
\newblock \URLprefix
  \url{https://campagnes.flotteoceanographique.fr/campagnes/17001000/},
  \DOIprefix\doi{10.17600/17001000}.
\bibitem[{Escart\'in et~al.(2016)Escart\'in, Leclerc, Olive, Mevel, Cannat,
  Petersen, Augustin, Feuillet, Deplus, Bezos, Bonnemains, Chavagnac, Choi,
  Godard, Haaga, Hamelin, Ildefonse, Jamieson, John, Leleu, Macleod,
  Massot-campos, Nomikou, Paquet, Rommevaux-Jestin, Rothenbeck, Steinfuhrer,
  Tominaga, Triebe, Campos, Gracias, Garcia, Andreani and
  Vilaseca}]{escartin2016}
\bibinfo{author}{Escart\'in, J.}, \bibinfo{author}{Leclerc, F.},
  \bibinfo{author}{Olive, J.A.}, \bibinfo{author}{Mevel, C.},
  \bibinfo{author}{Cannat, M.}, \bibinfo{author}{Petersen, S.},
  \bibinfo{author}{Augustin, N.}, \bibinfo{author}{Feuillet, N.},
  \bibinfo{author}{Deplus, C.}, \bibinfo{author}{Bezos, A.},
  \bibinfo{author}{Bonnemains, D.}, \bibinfo{author}{Chavagnac, V.},
  \bibinfo{author}{Choi, Y.}, \bibinfo{author}{Godard, M.},
  \bibinfo{author}{Haaga, K.}, \bibinfo{author}{Hamelin, C.},
  \bibinfo{author}{Ildefonse, B.}, \bibinfo{author}{Jamieson, J.W.},
  \bibinfo{author}{John, B.E.}, \bibinfo{author}{Leleu, T.},
  \bibinfo{author}{Macleod, C.J.}, \bibinfo{author}{Massot-campos, M.},
  \bibinfo{author}{Nomikou, P.}, \bibinfo{author}{Paquet, M.},
  \bibinfo{author}{Rommevaux-Jestin, C.}, \bibinfo{author}{Rothenbeck, M.},
  \bibinfo{author}{Steinfuhrer, A.}, \bibinfo{author}{Tominaga, M.},
  \bibinfo{author}{Triebe, L.}, \bibinfo{author}{Campos, R.},
  \bibinfo{author}{Gracias, N.}, \bibinfo{author}{Garcia, R.},
  \bibinfo{author}{Andreani, M.}, \bibinfo{author}{Vilaseca, G.},
  \bibinfo{year}{2016}.
\newblock \bibinfo{title}{First direct observation of coseismic slip and
  seafloor rupture along a submarine normal fault and implications for fault
  slip history}.
\newblock \bibinfo{journal}{Earth and Planetary Science Letters}
  \bibinfo{volume}{450}, \bibinfo{pages}{96--107}.
\newblock \DOIprefix\doi{10.1016/j.epsl.2016.06.024}.
\bibitem[{Escart{\'\i}n et~al.(2016)Escart{\'\i}n, Leclerc, Olive, Mevel,
  Cannat, Petersen, Augustin, Feuillet, Deplus, Bezos
  et~al.}]{escartin2016first}
\bibinfo{author}{Escart{\'\i}n, J.}, \bibinfo{author}{Leclerc, F.},
  \bibinfo{author}{Olive, J.A.}, \bibinfo{author}{Mevel, C.},
  \bibinfo{author}{Cannat, M.}, \bibinfo{author}{Petersen, S.},
  \bibinfo{author}{Augustin, N.}, \bibinfo{author}{Feuillet, N.},
  \bibinfo{author}{Deplus, C.}, \bibinfo{author}{Bezos, A.}, et~al.,
  \bibinfo{year}{2016}.
\newblock \bibinfo{title}{First direct observation of coseismic slip and
  seafloor rupture along a submarine normal fault and implications for fault
  slip history}.
\newblock \bibinfo{journal}{Earth and Planetary Science Letters}
  \bibinfo{volume}{450}, \bibinfo{pages}{96--107}.
\newblock \DOIprefix\doi{10.1016/j.epsl.2016.06.024}.
\bibitem[{Feuillet et~al.(2011)Feuillet, Beauducel, Jacques, Tapponnier,
  Delouis, Bazin, Vall{\'e}e and King}]{feuillet2011mw}
\bibinfo{author}{Feuillet, N.}, \bibinfo{author}{Beauducel, F.},
  \bibinfo{author}{Jacques, E.}, \bibinfo{author}{Tapponnier, P.},
  \bibinfo{author}{Delouis, B.}, \bibinfo{author}{Bazin, S.},
  \bibinfo{author}{Vall{\'e}e, M.}, \bibinfo{author}{King, G.},
  \bibinfo{year}{2011}.
\newblock \bibinfo{title}{The mw= 6.3, november 21, 2004, les saintes
  earthquake (guadeloupe): Tectonic setting, slip model and static stress
  changes}.
\newblock \bibinfo{journal}{Journal of Geophysical Research: Solid Earth}
  \bibinfo{volume}{116}.
\newblock \DOIprefix\doi{10.1029/2011JB008310}.
\bibitem[{Forlani et~al.(2018)Forlani, Dall'Asta, Diotri, Cella, Roncella and
  Santise}]{Forlani2018quality}
\bibinfo{author}{Forlani, G.}, \bibinfo{author}{Dall'Asta, E.},
  \bibinfo{author}{Diotri, F.}, \bibinfo{author}{Cella, U.M.d.},
  \bibinfo{author}{Roncella, R.}, \bibinfo{author}{Santise, M.},
  \bibinfo{year}{2018}.
\newblock \bibinfo{title}{Quality assessment of dsms produced from uav flights
  georeferenced with on-board rtk positioning}.
\newblock \bibinfo{journal}{Remote Sensing} \bibinfo{volume}{10}.
\newblock \DOIprefix\doi{10.3390/rs10020311}.
\bibitem[{Garcia et~al.(2011)Garcia, Campos and Escart\'in}]{garcia2011high}
\bibinfo{author}{Garcia, R.}, \bibinfo{author}{Campos, R.},
  \bibinfo{author}{Escart\'in, J.}, \bibinfo{year}{2011}.
\newblock \bibinfo{title}{High-resolution 3d reconstruction of the seafloor for
  environmental monitoring and modelling}, in: \bibinfo{booktitle}{Proc.
  Intelligent Robots and Systems (IROS), 2011 IEEE/RSJ International Conference
  on}.
\bibitem[{Hartley and Zisserman(2003)}]{zisserman2004multiple}
\bibinfo{author}{Hartley, R.}, \bibinfo{author}{Zisserman, A.},
  \bibinfo{year}{2003}.
\newblock \bibinfo{title}{Multiple View Geometry in Computer Vision}.
\newblock \bibinfo{edition}{2} ed., \bibinfo{publisher}{Cambridge University
  Press}, \bibinfo{address}{New York, NY, USA}.
\newblock \DOIprefix\doi{10.1017/CBO9780511811685.001}.
\bibitem[{Hern{\'a}ndez et~al.(2016)Hern{\'a}ndez, Isteni{\v{c}}, Gracias,
  Palomeras, Campos, Vidal, Garcia and Carreras}]{hernandez2016autonomous}
\bibinfo{author}{Hern{\'a}ndez, J.D.}, \bibinfo{author}{Isteni{\v{c}}, K.},
  \bibinfo{author}{Gracias, N.}, \bibinfo{author}{Palomeras, N.},
  \bibinfo{author}{Campos, R.}, \bibinfo{author}{Vidal, E.},
  \bibinfo{author}{Garcia, R.}, \bibinfo{author}{Carreras, M.},
  \bibinfo{year}{2016}.
\newblock \bibinfo{title}{Autonomous underwater navigation and optical mapping
  in unknown natural environments}.
\newblock \bibinfo{journal}{Sensors} \bibinfo{volume}{16},
  \bibinfo{pages}{1174}.
\newblock \DOIprefix\doi{10.3390/s16081174}.
\bibitem[{James and Robson(2014)}]{james2014mitigating}
\bibinfo{author}{James, M.R.}, \bibinfo{author}{Robson, S.},
  \bibinfo{year}{2014}.
\newblock \bibinfo{title}{Mitigating systematic error in topographic models
  derived from uav and ground-based image networks}.
\newblock \bibinfo{journal}{Earth Surface Processes and Landforms}
  \bibinfo{volume}{39}, \bibinfo{pages}{1413--1420}.
\newblock \DOIprefix\doi{10.1002/esp.3609}.
\bibitem[{Jancosek and Pajdla(2014)}]{jancosek2014exploiting}
\bibinfo{author}{Jancosek, M.}, \bibinfo{author}{Pajdla, T.},
  \bibinfo{year}{2014}.
\newblock \bibinfo{title}{Exploiting visibility information in surface
  reconstruction to preserve weakly supported surfaces}.
\newblock \bibinfo{journal}{International scholarly research notices}
  \bibinfo{volume}{2014}.
\newblock \DOIprefix\doi{10.1155/2014/798595}.
\bibitem[{Javernick et~al.(2014)Javernick, Brasington and
  Caruso}]{javernick2014modeling}
\bibinfo{author}{Javernick, L.}, \bibinfo{author}{Brasington, J.},
  \bibinfo{author}{Caruso, B.}, \bibinfo{year}{2014}.
\newblock \bibinfo{title}{Modeling the topography of shallow braided rivers
  using structure-from-motion photogrammetry}.
\newblock \bibinfo{journal}{Geomorphology} \bibinfo{volume}{213},
  \bibinfo{pages}{166--182}.
\newblock \DOIprefix\doi{10.1016/j.geomorph.2014.01.006}.
\bibitem[{Kalacska et~al.(2018)Kalacska, Lucanus, Sousa, Vieira and
  Arroyo-Mora}]{kalacska2018freshwater}
\bibinfo{author}{Kalacska, M.}, \bibinfo{author}{Lucanus, O.},
  \bibinfo{author}{Sousa, L.}, \bibinfo{author}{Vieira, T.},
  \bibinfo{author}{Arroyo-Mora, J.}, \bibinfo{year}{2018}.
\newblock \bibinfo{title}{Freshwater fish habitat complexity mapping using
  above and underwater structure-from-motion photogrammetry}.
\newblock \bibinfo{journal}{Remote Sensing} \bibinfo{volume}{10},
  \bibinfo{pages}{1912}.
\newblock \DOIprefix\doi{10.3390/rs10121912}.
\bibitem[{Ke and Roumeliotis(2017)}]{ke2017efficient}
\bibinfo{author}{Ke, T.}, \bibinfo{author}{Roumeliotis, S.I.},
  \bibinfo{year}{2017}.
\newblock \bibinfo{title}{An efficient algebraic solution to the
  perspective-three-point problem}, in: \bibinfo{booktitle}{Proceedings of the
  IEEE Conference on Computer Vision and Pattern Recognition}, pp.
  \bibinfo{pages}{7225--7233}.
\bibitem[{Kocak et~al.(2002)Kocak, Caimi, Jagielo and Kloske}]{kocak2002laser}
\bibinfo{author}{Kocak, D.M.}, \bibinfo{author}{Caimi, F.M.},
  \bibinfo{author}{Jagielo, T.H.}, \bibinfo{author}{Kloske, J.},
  \bibinfo{year}{2002}.
\newblock \bibinfo{title}{Laser projection photogrammetry and video system for
  quantification and mensuration}, in: \bibinfo{booktitle}{OCEANS '02
  MTS/IEEE}, pp. \bibinfo{pages}{1569--1574 vol.3}.
\newblock \DOIprefix\doi{10.1109/OCEANS.2002.1191869}.
\bibitem[{Kocak et~al.(2004)Kocak, Jagielo, Wallace and
  Kloske}]{kocak2004remote}
\bibinfo{author}{Kocak, D.M.}, \bibinfo{author}{Jagielo, T.H.},
  \bibinfo{author}{Wallace, F.}, \bibinfo{author}{Kloske, J.},
  \bibinfo{year}{2004}.
\newblock \bibinfo{title}{Remote sensing using laser projection photogrammetry
  for underwater surveys}, in: \bibinfo{booktitle}{IGARSS 2004. 2004 IEEE
  International Geoscience and Remote Sensing Symposium}, pp.
  \bibinfo{pages}{1451--1454 vol.2}.
\newblock \DOIprefix\doi{10.1109/IGARSS.2004.1368693}.
\bibitem[{Lourakis and Zabulis(2013)}]{lourakis2013accurate}
\bibinfo{author}{Lourakis, M.}, \bibinfo{author}{Zabulis, X.},
  \bibinfo{year}{2013}.
\newblock \bibinfo{title}{Accurate scale factor estimation in 3d
  reconstruction}, in: \bibinfo{booktitle}{International Conference on Computer
  Analysis of Images and Patterns}, \bibinfo{organization}{Springer}. pp.
  \bibinfo{pages}{498--506}.
\bibitem[{Mathews and Jensen(2013)}]{mathews2013}
\bibinfo{author}{Mathews, A.J.}, \bibinfo{author}{Jensen, J.L.R.},
  \bibinfo{year}{2013}.
\newblock \bibinfo{title}{Visualizing and quantifying vineyard canopy lai using
  an unmanned aerial vehicle (uav) collected high density structure from motion
  point cloud}.
\newblock \bibinfo{journal}{Remote Sensing} \bibinfo{volume}{5},
  \bibinfo{pages}{2164--2183}.
\newblock \DOIprefix\doi{10.3390/rs5052164}.
\bibitem[{Mertes et~al.(2017)Mertes, Zant, Gulley and
  Thomsen}]{mertes2017rapid}
\bibinfo{author}{Mertes, J.}, \bibinfo{author}{Zant, C.},
  \bibinfo{author}{Gulley, J.}, \bibinfo{author}{Thomsen, T.},
  \bibinfo{year}{2017}.
\newblock \bibinfo{title}{Rapid, quantitative assessment of submerged cultural
  resource degradation using repeat video surveys and structure from motion}.
\newblock \bibinfo{journal}{Journal of Maritime Archaeology}
  \bibinfo{volume}{12}, \bibinfo{pages}{91--107}.
\newblock \DOIprefix\doi{10.1007/s11457-017-9172-0}.
\bibitem[{Mian et~al.(2016)Mian, Lutes, Lipa, Hutton, Gavelle and
  Borghini}]{mian2016accuracy}
\bibinfo{author}{Mian, O.}, \bibinfo{author}{Lutes, J.}, \bibinfo{author}{Lipa,
  G.}, \bibinfo{author}{Hutton, J.}, \bibinfo{author}{Gavelle, E.},
  \bibinfo{author}{Borghini, S.}, \bibinfo{year}{2016}.
\newblock \bibinfo{title}{Accuracy assessment of direct georeferencing for
  photogrammetric applications on small unmanned aerial platforms}.
\newblock \bibinfo{journal}{The International Archives of Photogrammetry,
  Remote Sensing and Spatial Information Sciences} \bibinfo{volume}{40},
  \bibinfo{pages}{77}.
\newblock \DOIprefix\doi{10.5194/isprs-archives-XL-3-W4-77-2016}.
\bibitem[{Michel et~al.(2003)Michel, Klages, Barriga, Fouquet, Sibuet,
  Sarradin, Sim{\'e}oni, Drogou et~al.}]{Michel2003}
\bibinfo{author}{Michel, J.L.}, \bibinfo{author}{Klages, M.},
  \bibinfo{author}{Barriga, F.J.}, \bibinfo{author}{Fouquet, Y.},
  \bibinfo{author}{Sibuet, M.}, \bibinfo{author}{Sarradin, P.M.},
  \bibinfo{author}{Sim{\'e}oni, P.}, \bibinfo{author}{Drogou, J.F.}, et~al.,
  \bibinfo{year}{2003}.
\newblock \bibinfo{title}{Victor 6000: design, utilization and first
  improvements}, in: \bibinfo{booktitle}{The Thirteenth International Offshore
  and Polar Engineering Conference}, \bibinfo{organization}{International
  Society of Offshore and Polar Engineers}.
\bibitem[{Moisan et~al.(2012)Moisan, Moulon and Monasse}]{moisan2012automatic}
\bibinfo{author}{Moisan, L.}, \bibinfo{author}{Moulon, P.},
  \bibinfo{author}{Monasse, P.}, \bibinfo{year}{2012}.
\newblock \bibinfo{title}{Automatic homographic registration of a pair of
  images, with a contrario elimination of outliers}.
\newblock \bibinfo{journal}{Image Processing On Line} \bibinfo{volume}{2},
  \bibinfo{pages}{56--73}.
\newblock \DOIprefix\doi{10.5201/ipol.2012.mmm-oh}.
\bibitem[{Moulon et~al.(2013)Moulon, Monasse and Marlet}]{moulon2013global}
\bibinfo{author}{Moulon, P.}, \bibinfo{author}{Monasse, P.},
  \bibinfo{author}{Marlet, R.}, \bibinfo{year}{2013}.
\newblock \bibinfo{title}{Global fusion of relative motions for robust,
  accurate and scalable structure from motion}, in:
  \bibinfo{booktitle}{Proceedings of the IEEE International Conference on
  Computer Vision}, pp. \bibinfo{pages}{3248--3255}.
\newblock \DOIprefix\doi{10.1109/ICCV.2013.403}.
\bibitem[{Moulon et~al.()Moulon, Monasse, Marlet and Others}]{openMVG}
\bibinfo{author}{Moulon, P.}, \bibinfo{author}{Monasse, P.},
  \bibinfo{author}{Marlet, R.}, \bibinfo{author}{Others}, .
\newblock \bibinfo{title}{Openmvg. an open multiple view geometry library.}
\newblock \bibinfo{howpublished}{\url{https://github.com/openMVG/openMVG}}.
\bibitem[{Neyer et~al.(2018)Neyer, Nocerino and Gruen}]{neyer2018monitoring}
\bibinfo{author}{Neyer, F.}, \bibinfo{author}{Nocerino, E.},
  \bibinfo{author}{Gruen, A.}, \bibinfo{year}{2018}.
\newblock \bibinfo{title}{Monitoring coral growth-the dichotomy between
  underwater photogrammetry and geodetic control network.}
\newblock \bibinfo{journal}{International Archives of the Photogrammetry,
  Remote Sensing and Spatial Information Sciences} \bibinfo{volume}{42},
  \bibinfo{pages}{2}.
\newblock \DOIprefix\doi{10.5194/isprs-archives-XLII-2-759-2018}.
\bibitem[{Pilgrim et~al.(2000)Pilgrim, Parry, Jones and
  Kendall}]{pilgrim2000rov}
\bibinfo{author}{Pilgrim, D.A.}, \bibinfo{author}{Parry, D.M.},
  \bibinfo{author}{Jones, M.B.}, \bibinfo{author}{Kendall, M.A.},
  \bibinfo{year}{2000}.
\newblock \bibinfo{title}{Rov image scaling with laser spot patterns}.
\newblock \bibinfo{journal}{Underwater Technology} \bibinfo{volume}{24},
  \bibinfo{pages}{93--103}.
\newblock \DOIprefix\doi{10.3723/175605400783259684}.
\bibitem[{Pizarro et~al.(2009)Pizarro, Eustice and Singh}]{pizarro2009large}
\bibinfo{author}{Pizarro, O.}, \bibinfo{author}{Eustice, R.M.},
  \bibinfo{author}{Singh, H.}, \bibinfo{year}{2009}.
\newblock \bibinfo{title}{Large area 3-d reconstructions from underwater
  optical surveys}.
\newblock \bibinfo{journal}{IEEE Journal of Oceanic Engineering}
  \bibinfo{volume}{34}, \bibinfo{pages}{150--169}.
\newblock \DOIprefix\doi{10.1109/JOE.2009.2016071}.
\bibitem[{Pizarro et~al.(2017)Pizarro, Friedman, Bryson, Williams and
  Madin}]{pizarro2017simple}
\bibinfo{author}{Pizarro, O.}, \bibinfo{author}{Friedman, A.},
  \bibinfo{author}{Bryson, M.}, \bibinfo{author}{Williams, S.B.},
  \bibinfo{author}{Madin, J.}, \bibinfo{year}{2017}.
\newblock \bibinfo{title}{A simple, fast, and repeatable survey method for
  underwater visual 3d benthic mapping and monitoring}.
\newblock \bibinfo{journal}{Ecology and Evolution} \bibinfo{volume}{7},
  \bibinfo{pages}{1770--1782}.
\newblock \DOIprefix\doi{10.1002/ece3.2701}.
\bibitem[{Remondino et~al.(2008)Remondino, El-Hakim, Gruen and
  Zhang}]{remondino2008turning}
\bibinfo{author}{Remondino, F.}, \bibinfo{author}{El-Hakim, S.F.},
  \bibinfo{author}{Gruen, A.}, \bibinfo{author}{Zhang, L.},
  \bibinfo{year}{2008}.
\newblock \bibinfo{title}{Turning images into 3-d models}.
\newblock \bibinfo{journal}{IEEE Signal Processing Magazine}
  \bibinfo{volume}{25}, \bibinfo{pages}{55--65}.
\newblock \DOIprefix\doi{10.1109/MSP.2008.923093}.
\bibitem[{Robert et~al.(2017)Robert, Huvenne, Georgiopoulou, Jones, Marsh,
  Carter and Chaumillon}]{robert2017new}
\bibinfo{author}{Robert, K.}, \bibinfo{author}{Huvenne, V.A.},
  \bibinfo{author}{Georgiopoulou, A.}, \bibinfo{author}{Jones, D.O.},
  \bibinfo{author}{Marsh, L.}, \bibinfo{author}{Carter, G.D.},
  \bibinfo{author}{Chaumillon, L.}, \bibinfo{year}{2017}.
\newblock \bibinfo{title}{New approaches to high-resolution mapping of marine
  vertical structures}.
\newblock \bibinfo{journal}{Scientific reports} \bibinfo{volume}{7},
  \bibinfo{pages}{9005}.
\bibitem[{Rossi et~al.(2019)Rossi, Castagnetti, Capra, Brooks and
  Mancini}]{rossi2019detecting}
\bibinfo{author}{Rossi, P.}, \bibinfo{author}{Castagnetti, C.},
  \bibinfo{author}{Capra, A.}, \bibinfo{author}{Brooks, A.},
  \bibinfo{author}{Mancini, F.}, \bibinfo{year}{2019}.
\newblock \bibinfo{title}{Detecting change in coral reef 3d structure using
  underwater photogrammetry: critical issues and performance metrics}.
\newblock \bibinfo{journal}{Applied Geomatics} ,
  \bibinfo{pages}{1--15}\DOIprefix\doi{10.1007/s12518-019-00263-w}.
\bibitem[{{Rzhanov} et~al.(2005){Rzhanov}, {Mamaenko} and
  {Yoklavich}}]{rzhanov2005uvsd}
\bibinfo{author}{{Rzhanov}, Y.}, \bibinfo{author}{{Mamaenko}, A.},
  \bibinfo{author}{{Yoklavich}, M.}, \bibinfo{year}{2005}.
\newblock \bibinfo{title}{Uvsd: software for detection of color underwater
  features}, in: \bibinfo{booktitle}{Proceedings of OCEANS 2005 MTS/IEEE}, pp.
  \bibinfo{pages}{2189--2192 Vol. 3}.
\newblock \DOIprefix\doi{10.1109/OCEANS.2005.1640089}.
\bibitem[{Sedlazeck et~al.(2009)Sedlazeck, Koser and Koch}]{sedlazeck2009kiel}
\bibinfo{author}{Sedlazeck, A.}, \bibinfo{author}{Koser, K.},
  \bibinfo{author}{Koch, R.}, \bibinfo{year}{2009}.
\newblock \bibinfo{title}{3d reconstruction based on underwater video from rov
  kiel 6000 considering underwater imaging conditions}, in:
  \bibinfo{booktitle}{OCEANS 2009-EUROPE}, pp. \bibinfo{pages}{1--10}.
\newblock \DOIprefix\doi{10.1109/OCEANSE.2009.5278305}.
\bibitem[{Shen(2013)}]{shen2013accurate}
\bibinfo{author}{Shen, S.}, \bibinfo{year}{2013}.
\newblock \bibinfo{title}{Accurate multiple view 3d reconstruction using
  patch-based stereo for large-scale scenes}.
\newblock \bibinfo{journal}{IEEE transactions on image processing}
  \bibinfo{volume}{22}, \bibinfo{pages}{1901--1914}.
\newblock \DOIprefix\doi{10.1109/TIP.2013.2237921}.
\bibitem[{Snavely et~al.(2008)Snavely, Seitz and
  Szeliski}]{snavely2008modeling}
\bibinfo{author}{Snavely, N.}, \bibinfo{author}{Seitz, S.M.},
  \bibinfo{author}{Szeliski, R.}, \bibinfo{year}{2008}.
\newblock \bibinfo{title}{Modeling the world from internet photo collections}.
\newblock \bibinfo{journal}{International Journal of Computer Vision}
  \bibinfo{volume}{80}, \bibinfo{pages}{189--210}.
\newblock \DOIprefix\doi{10.1007/s11263-007-0107-3}.
\bibitem[{Soloviev and Venable(2010)}]{soloviev2010integration}
\bibinfo{author}{Soloviev, A.}, \bibinfo{author}{Venable, D.},
  \bibinfo{year}{2010}.
\newblock \bibinfo{title}{Integration of gps and vision measurements for
  navigation in gps challenged environments}, in: \bibinfo{booktitle}{IEEE/ION
  Position, Location and Navigation Symposium}, pp. \bibinfo{pages}{826--833}.
\newblock \DOIprefix\doi{10.1109/PLANS.2010.5507322}.
\bibitem[{Spaenlehauer et~al.(2017)Spaenlehauer, Fremont, Sekercioglu and
  Fantoni}]{spaenlehauer2017loosely}
\bibinfo{author}{Spaenlehauer, A.}, \bibinfo{author}{Fremont, V.},
  \bibinfo{author}{Sekercioglu, Y.A.}, \bibinfo{author}{Fantoni, I.},
  \bibinfo{year}{2017}.
\newblock \bibinfo{title}{A loosely-coupled approach for metric scale
  estimation in monocular vision-inertial systems}, in:
  \bibinfo{booktitle}{2017 IEEE International Conference on Multisensor Fusion
  and Integration for Intelligent Systems (MFI)}, pp.
  \bibinfo{pages}{137--143}.
\newblock \DOIprefix\doi{10.1109/MFI.2017.8170419}.
\bibitem[{Storlazzi et~al.(2016)Storlazzi, Dartnell, Hatcher and
  Gibbs}]{storlazzi2016end}
\bibinfo{author}{Storlazzi, C.D.}, \bibinfo{author}{Dartnell, P.},
  \bibinfo{author}{Hatcher, G.A.}, \bibinfo{author}{Gibbs, A.E.},
  \bibinfo{year}{2016}.
\newblock \bibinfo{title}{End of the chain? rugosity and fine-scale bathymetry
  from existing underwater digital imagery using structure-from-motion (sfm)
  technology}.
\newblock \bibinfo{journal}{Coral Reefs} \bibinfo{volume}{35},
  \bibinfo{pages}{889--894}.
\newblock \DOIprefix\doi{10.1007/s00338-016-1462-8}.
\bibitem[{Triggs et~al.(1999)Triggs, McLauchlan, Hartley and
  Fitzgibbon}]{triggs1999bundle}
\bibinfo{author}{Triggs, B.}, \bibinfo{author}{McLauchlan, P.F.},
  \bibinfo{author}{Hartley, R.I.}, \bibinfo{author}{Fitzgibbon, A.W.},
  \bibinfo{year}{1999}.
\newblock \bibinfo{title}{Bundle adjustment -- a modern synthesis}, in:
  \bibinfo{booktitle}{Vision algorithms: theory and practice}.
  \bibinfo{publisher}{Springer}, pp. \bibinfo{pages}{298--372}.
\newblock \DOIprefix\doi{10.1007/3-540-44480-7_21}.
\bibitem[{Tusting and Davis(1986)}]{tusting1986noncon}
\bibinfo{author}{Tusting, R.}, \bibinfo{author}{Davis, D.},
  \bibinfo{year}{1986}.
\newblock \bibinfo{title}{Non-conventional techniques for sampling and
  collecting marine organisms}, in: \bibinfo{booktitle}{Proceedings of the
  Pacific Congress on Marine Technology, PACON'86}, pp.
  \bibinfo{pages}{12--18}.
\bibitem[{Tusting and Davis(1993)}]{tusting1993improved}
\bibinfo{author}{Tusting, R.F.}, \bibinfo{author}{Davis, D.},
  \bibinfo{year}{1993}.
\newblock \bibinfo{title}{Improved methods for visual and photographic benthic
  surveys} .
\bibitem[{Tusting and Davis(1992)}]{tusting1992laser}
\bibinfo{author}{Tusting, R.F.}, \bibinfo{author}{Davis, D.L.},
  \bibinfo{year}{1992}.
\newblock \bibinfo{title}{Laser systems and structured illumination for
  quantitative undersea imaging}.
\newblock \bibinfo{journal}{Marine Technology Society Journal}
  \bibinfo{volume}{26}, \bibinfo{pages}{5--12}.
\bibitem[{Waechter et~al.(2014)Waechter, Moehrle and Goesele}]{waechter2014let}
\bibinfo{author}{Waechter, M.}, \bibinfo{author}{Moehrle, N.},
  \bibinfo{author}{Goesele, M.}, \bibinfo{year}{2014}.
\newblock \bibinfo{title}{Let there be color! large-scale texturing of 3d
  reconstructions}, in: \bibinfo{booktitle}{Computer Vision--ECCV}.
  \bibinfo{publisher}{Springer}, pp. \bibinfo{pages}{836--850}.
\newblock \DOIprefix\doi{10.1007/978-3-319-10602-1_54}.
\bibitem[{Wakefield and Genin(1987)}]{wakefield1987canadian}
\bibinfo{author}{Wakefield, W.W.}, \bibinfo{author}{Genin, A.},
  \bibinfo{year}{1987}.
\newblock \bibinfo{title}{The use of a canadian (perspective) grid in deep-sea
  photography}.
\newblock \bibinfo{journal}{Deep Sea Research Part A. Oceanographic Research
  Papers} \bibinfo{volume}{34}, \bibinfo{pages}{469 -- 478}.
\newblock \DOIprefix\doi{10.1016/0198-0149(87)90148-8}.
\bibitem[{Wallace et~al.(2016)Wallace, Lucieer, Malenovsky, Turner and
  Vopenka}]{wallace2016assessment}
\bibinfo{author}{Wallace, L.}, \bibinfo{author}{Lucieer, A.},
  \bibinfo{author}{Malenovsky, Z.}, \bibinfo{author}{Turner, D.},
  \bibinfo{author}{Vopenka, P.}, \bibinfo{year}{2016}.
\newblock \bibinfo{title}{Assessment of forest structure using two uav
  techniques: A comparison of airborne laser scanning and structure from motion
  (sfm) point clouds}.
\newblock \bibinfo{journal}{Forests} \bibinfo{volume}{7}.
\newblock \DOIprefix\doi{10.3390/f7030062}.
\bibitem[{Zhang and Singh(2015)}]{ji2015visual}
\bibinfo{author}{Zhang, J.}, \bibinfo{author}{Singh, S.}, \bibinfo{year}{2015}.
\newblock \bibinfo{title}{Visual-inertial combined odometry system for aerial
  vehicles}.
\newblock \bibinfo{journal}{Journal of Field Robotics} \bibinfo{volume}{32},
  \bibinfo{pages}{1043--1055}.
\newblock \DOIprefix\doi{10.1002/rob.21599}.

\end{thebibliography}


\bio[width=10mm,pos=l]{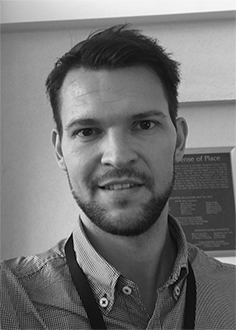}
 \textbf{Klemen Isteni\v{c}:}
  received his Diploma degree in Computer Science from the University of Ljubljana, Slovenia in 2013 and a joint M.Sc. degree with distinction in Computer Vision and Robotics (European Master ViBOT) from Heriott-Watt University, UK, University of Girona, Spain, and the University of Burgundy, France in 2015. He is currently pursuing a Ph.D. in Technology at the University of Girona as a member of the Underwater Robotics Research Center (CIRS), part of the Computer Vision and Robotics Institute, as well as a member of an European Academy for Marine and Underwater Robotics (EU FP7 Marie Curie ITN network no 608096 - Robocademy). His research focuses on 3D mapping, color restoration and change detection using optical data in underwater scenarios.
\endbio

\bio[width=10mm,pos=l]{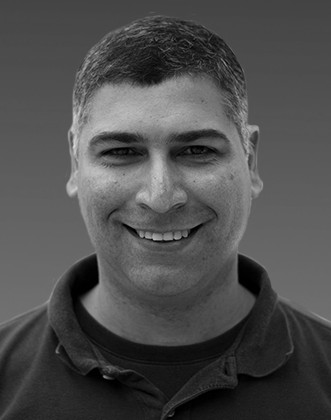}
 \textbf{Dr. Nuno Gracias:}
  was awarded the Ph.D. degree in 2003 from the Technical University of Lisbon, Portugal. From 2004 to 2006 he was a post-doctorate fellow at the University of Miami. Since 2006 he has been a member of the Computer Vision and Robotics Group (ViCOROB) of the University of Girona. His research interests span the areas of underwater optical mapping, and navigation and guidance of autonomous underwater robots, image processing and classification. Dr Gracias has authored more than 80 articles in peer-review journals and scientific conferences, and co-supervised 3 PhD and 8 MSc theses. He is adjunct faculty at the department of Marine Geosciences of the University of Miami, and member of the editorial board of the Journal of Intelligent and Robotic Systems.
\endbio

\bio[width=10mm,pos=l]{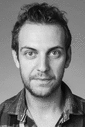}
 \textbf{Dr. Aur\'elien Arnaubec:} received his Ph.D degree from the Univers\'e Aix-Marseille III,Marseille, France, in 2012. He did his Ph.D. degree at the French Aerospace Laboratory, Office National d'Etudes et Recherches A\'erospatiales, (ONERA), Salon Cedex Air, France, and in the Physics and Image Processing Group, Fresnel Institute, Marseille, France where his main research interests  was radar imaging and statistical signal processing for remote sensing. Then from 2012 to 2014 he did a postdoc at Ifremer, la Seyne sur Mer, France, where he developped image processing techniques for optical mapping. He now joined Ifremer PRAO team (Positionning, Robotic, Acoustic and Optics ) where he is in charge of all underwater optical systems and image processing softwares.
 \endbio

\bio[width=10mm,pos=l]{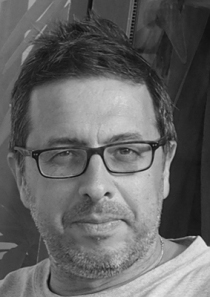}
 \textbf{Dr. Javier Escart\'in:} received his PhD from the MIT/WHOI Joint Program in 1996 (USA). He is now a CNRS Senior Research Scientist at the Institute de Physique du Globe de Paris - Universit\'e de Paris, where he leads since 2017 the Marine Geosciences Group. His research focuses in deep-sea exploration of the seafloor to understand geological processes, such as tectonism, volcanism, or hydrothermal activity. For this research he uses acoustic and optical mapping of the seafloor acquired with deep-sea vehicles.
\endbio

\bio[width=10mm,pos=l]{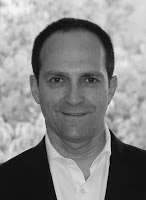}
 \textbf{Dr. Rafael Garcia:}
  received a M.S. degree in computer engineering from the Universitat Autonoma de Barcelona in 1994, and a Ph.D. in Computer Engineering from the Universitat de Girona (Spain) in 2001. He is the founder and director of the Underwater Vision Lab (UVL), within the Computer Vision and Robotics Group in the Department of Computer Architecture at the University of Girona. His main research interests are underwater robotics and computer vision. He is particularly interested in how to make underwater vehicles sense their environment in order to carry out autonomous surveys. He has published more than 190 technical contributions, including journal papers and conference proceedings. Dr. Garcia is a Member of the IEEE.
\endbio

\end{document}